\documentclass[lettersize,journal]{IEEEtran}
\usepackage{amsmath,amsfonts}
\usepackage{algorithm}
\usepackage{array}
\usepackage[caption=false,font=normalsize,labelfont=sf,textfont=sf]{subfig}
\usepackage{textcomp}
\usepackage{stfloats}
\usepackage{url}
\usepackage{verbatim}
\usepackage{graphicx}
\usepackage[utf8]{inputenc}         
\usepackage[T1]{fontenc}            
\usepackage{hyperref}               
\usepackage{nicefrac}               
\usepackage{microtype}              
\usepackage{amsfonts}               
\usepackage{amsmath}                
\usepackage[dvipsnames]{xcolor}     
\usepackage{float}                  
\graphicspath{{images/}}
\DeclareGraphicsExtensions{.pdf,.PDF,.jpg,.JPG,.jpeg,.JPEG,.png,.PNG}
\usepackage{booktabs}               
\usepackage{multicol}               
\usepackage{multirow}               
\newcolumntype{C}[1]{>{\centering\arraybackslash}p{#1}}

\usepackage[nolist]{acronym}        
\usepackage{comment}                

\usepackage[numbers]{natbib}
\usepackage[normalem]{ulem}
\usepackage{diagbox}
\usepackage{makecell}
\usepackage{algpseudocode}
\usepackage[super]{nth}            
\usepackage{arydshln}   
\usepackage{multirow} 

\newcommand{\etal}{\textit{et al}. }
\newcommand{\ie}{\textit{i}.\textit{e}., }

\usepackage{xcolor}

\newcommand\tocite[1]{\textcolor{blue}{[REFERENCE]}}

\begin{document}

\title{A Systematic Performance Analysis of Deep Perceptual Loss Networks:\\ \huge{Breaking Transfer Learning Conventions}}

\author{
    \IEEEauthorblockN{
        \textbf{Gustav~Grund~Pihlgren}\IEEEauthorrefmark{1}\IEEEauthorrefmark{2}, \and
        \textbf{Konstantina~Nikolaidou}\IEEEauthorrefmark{2}, \and
        \textbf{Prakash~Chandra~Chhipa}\IEEEauthorrefmark{2}, \and
        \textbf{Nosheen~Abid}\IEEEauthorrefmark{2}, \and
        \textbf{Rajkumar~Saini}\IEEEauthorrefmark{2}, \and
        \textbf{Fredrik~Sandin}\IEEEauthorrefmark{2}, \and
        \textbf{Marcus~Liwicki}\IEEEauthorrefmark{2}
    }\\
    \vspace{0.2cm}
    \IEEEauthorblockA{
        \IEEEauthorrefmark{1}%
        \textit{Explainable AI Group}\\
        Umeå University, Sweden\\
        gustav.pihlgren@umu.se\\
        \vspace{0.15cm}
        \IEEEauthorrefmark{2}
        \textit{Machine Learning Group}\\
        Lule{\aa} University of Technology, Sweden\\
        \{firstname\}.\{lastname\}@ltu.se\\
    }
}




\maketitle

\begin{abstract}

In recent years, deep perceptual loss has been widely and successfully used to train machine learning models for many computer vision tasks, including image synthesis, segmentation, and autoencoding.
Deep perceptual loss is a type of loss function for images that computes the error between two images as the distance between deep features extracted from a neural network.
Most applications of the loss use pretrained networks called loss networks for deep feature extraction.
However, despite increasingly widespread use, the effects of loss network implementation on the trained models have not been studied.


This work rectifies this through a systematic evaluation of the effect of different pretrained loss networks on four different application areas.
Specifically, the work evaluates 14 different pretrained architectures with four different feature extraction layers.
The evaluation reveals that VGG networks without batch normalization have the best performance and that the choice of feature extraction layer is at least as important as the choice of architecture.
The analysis also reveals that deep perceptual loss does not adhere to the transfer learning conventions that better ImageNet accuracy implies better downstream performance and that feature extraction from the later layers provides better performance.

\end{abstract}


\section{Introduction}
\label{toc:introduction}





The calculation of loss functions is currently an essential part of training the most successful machine learning models for computer vision.
The ability to calculate the difference between a model output and the target output is, in turn, an essential component of most loss functions.
For models with image outputs, this comparison has traditionally been performed by comparing the individual pixels of the output image and the target image, so-called pixel-wise loss.
However, pixel-wise losses have flaws, such as disregarding inter-pixel dependencies and weighing all regions of the image equally~\cite{pihlgren2020improving}.

One successful direction of research into improved loss calculation is to use an auxiliary neural network as part of the loss calculation of the one being trained.
Methods following this direction include milestone achievements such as adversarial examples~\cite{szegedy2014intriguing}, generative adversarial networks \cite{goodfellow2014generative}, and deep visualization~\cite{yosinski2015understanding}.
The auxiliary neural networks used for loss calculation in these cases are called loss networks.


\begin{figure}[t] 
    \includegraphics[width=\columnwidth]{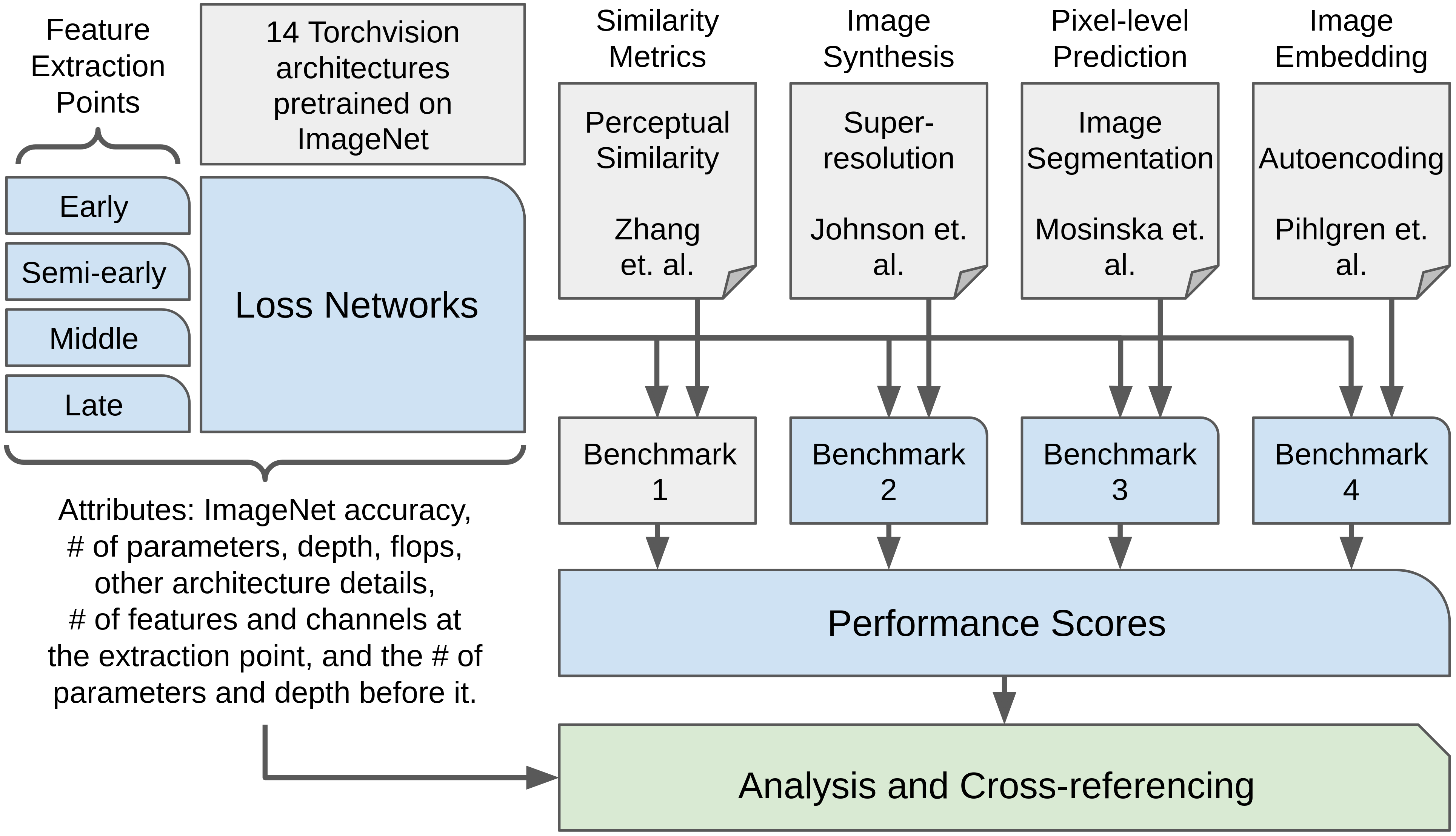}
    \centering
    \caption{
        The procedure followed in this work.
        This work investigates the effect of loss networks with different ImageNet~\cite{deng2009imagenet} pretrained architectures and feature extraction layers on the downstream performance of deep perceptual loss and similarity applications.
        Loss networks with $14$ pretrained architectures are examined for four different feature extraction layers by evaluating them on four application areas of deep perceptual loss and similarity.
        For each application area, a benchmark derived from a prior work is used for evaluation~\cite{zhang2018unreasonable, johnson2016perceptual, mosinska2018beyond, pihlgren2021pretraining}.
        The attributes and performance of each loss network are analyzed and cross-referenced with the other loss networks to uncover which attributes are correlated with performance and other trends.
        This work makes novel contributions (blue, round-corner) regarding feature extraction layers, systematic analysis of loss networks, and systematic evaluation on Benchmarks~2 through~4.
        The major contribution of this work (green, cut-corner) is the large analysis and cross-reference of the attributes and performance scores.
    }
    \label{fig:visual_abstract}
\end{figure}


A popular method that makes use of loss networks that has proven effective for a range of computer vision applications is deep perceptual loss.
Deep perceptual loss aims to create loss functions for machine learning models that mimic human perception of similarity.
This is typically done by calculating the similarity between the output image and ground truth by measuring how similar the deep features (activations) of the loss network are when each image is used as input.
This method of similarity measurement is known as deep perceptual similarity, and when used as part of a loss function it is known as deep perceptual loss.



Deep perceptual loss is well suited for image synthesis, where it is utilized to improve performance tasks, such as style-transfer~\cite{gatys2016image}, image generation~\cite{larsen2016autoencoding}, super-resolution~\cite{johnson2016perceptual}, and image denoising~\cite{yang2018low}.
In addition to image synthesis, the method has been successful for tasks with image-like outputs such as image segmentation~\cite{chai2020perceptual}, dimensionality reduction~\cite{pihlgren2020improving}, and image depth prediction~\cite{liu2021perceptual}.
Additionally, deep perceptual similarity has become one of the dominant techniques for measuring the perceptual similarity of images~\cite{zhang2018unreasonable, ding2020image}.

Despite the diverse and widespread use of deep perceptual loss and similarity, how to implement a suitable loss network for a given task remains unexplored.
The choice of loss network architectures used in prior works has either been justified by prior use~\cite{pihlgren2020improving, liu2021perceptual} or not at all~\cite{johnson2016perceptual,mosinska2018beyond,pihlgren2021pretraining,gatys2016image,chai2020perceptual,dosovitskiy2016generating}.
However, some rare exceptions give justifications based on hypotheses regarding the suitability of the architecture to the given data~\cite{xu2020infrared}, though these hypotheses are not actually tested.
Furthermore, the choice of extraction layers also typically lacks justification~\cite{johnson2016perceptual, pihlgren2021pretraining, chai2020perceptual, liu2021perceptual}.
In the cases when a justification does exist it is usually not tested~\cite{pihlgren2020improving, gatys2016image, dosovitskiy2016generating} or limited to similar layers~\cite{mosinska2018beyond}.
With the increasing use of deep perceptual loss, a study into improved loss network selection is well justified.


The aim of this work is to provide a systematic evaluation of loss networks in order to provide insight into which loss network implementations are most suitable for a given task.
Specifically, this work focuses on Convolutional Neural Network (CNN) architectures pretrained in ImageNet.
The work evaluates 14 different architectures with deep features extracted at four different layers on four application areas.
The loss networks are evaluated on one benchmark for each application area.
The performance scores gathered from the evaluation on the benchmarks are analyzed to provide insight into what loss network attributes are correlated with performance under different conditions.
An illustration of the procedure followed in this work, from the four prior works and loss networks to the analysis, is shown in Fig~\ref{fig:visual_abstract} and the complete implementation is available online\footnote{\url{https://github.com/LTU-Machine-Learning/Analysis-of-Deep-Perceptual-Loss-Networks}}.
Additional supplementary materials are described in the Appendix.

\subsection{Scope}
There are many aspects to consider when implementing a loss network for deep perceptual loss.
There is an abundance of neural network architectures available, plenty of datasets that could be used to train them, and a plethora of ways to perform feature extraction from those networks.
It is clearly outside the scope of any one study to look into them all, but through analyzing carefully curated combinations much insight can be gained.

This work investigates how the selection of pretrained architecture and feature extraction layers affects the performance of loss networks when used on different deep perceptual loss tasks.
It does this by systematically evaluating loss networks based on 14 easily accessible and well-known pretrained CNN architectures and comparing their performance as loss networks with deep features extracted at four different layers.
The loss networks are evaluated as perceptual similarity metrics and as part of deep perceptual loss to train models.
For deep perceptual similarity, they are applied as metrics of similarity and scored by their alignment with human perception.
For deep perceptual loss, they are applied to train models for each of the three major application areas of deep perceptual loss; image synthesis and transformation, pixel-level prediction, and image embedding.

For each of the four application areas, a benchmark task based on prior work is used for evaluation.
The benchmark tasks used in this work are perceptual similarity measurement~\cite{zhang2018unreasonable}, training image transformation networks for super-resolution~\cite{johnson2016perceptual}, training U-nets~\cite{ronneberger2015u} for image segmentation~\cite{mosinska2018beyond}, and training autoencoders for embeddings used in downstream prediction~\cite{pihlgren2021pretraining}.
The results obtained from the benchmark tasks are analyzed to gain insight into how the choice of architecture and feature extraction layer affects performance.
These insights have been the basis for further suggestions on how to select pretrained architectures and extraction layers for loss networks.
Details regarding the four benchmarks and justification for their use can be found in Section~\ref{sec:benchmarks}.

This work focuses on the use of pretrained networks as they are by far the most common use of deep perceptual loss and are easier to implement.
The pretrained models selected for evaluation are some of the classification architectures provided by Torchvision~\cite{marcel2010torchvision}, a package of the PyTorch~\cite{paszke2019pytorch} framework.
All of the selected models have been pretrained with the ImageNet~\cite{deng2009imagenet} dataset.
These networks, with or without pretrained parameters, are readily available for public use, which is itself a motivation for their usage; however, there are more reasons for using the networks provided in Torchvision.
The pretrained neural networks available in Torchvision contain a wide variety of architectures based on several renowned papers~\cite{krizhevsky2012imagenet,krizhevsky2014one,simonyan2015very,he2016deep,szegedy2015going}.
There are also multiple different implementations of the different networks within Torchvision.
This provides the ability to test how different architectures behave as loss networks and how this behavior changes with more minor changes in implementation.
Additionally, applications of deep perceptual loss commonly use the pretrained networks in Torchvision to create loss networks.

This work only considers models pretrained on ImageNet in the loss networks.
Aside from ease of access, the ImageNet dataset is a good choice for loss network pretraining as it is the most popular pretraining dataset for deep perceptual loss.
Additionally, it is one of the largest and most well-used computer vision datasets.

This work also investigates how the choice of layers in the loss networks from which features are extracted affects the trained models.
CNNs typically have millions of parameters across several layers from which features could be extracted.
Therefore this work focuses on whether to extract from the earlier, mid, or later layers.
The impact of different layers for feature extraction has often been left out of prior works and no systematic evaluation has been conducted.

In this work, the loss networks extract features in a straightforward manner.
For a given architecture and extraction layer, the activations are propagated forward to the extraction layer and then used directly to calculate the loss, as detailed in Section~\ref{sec:loss_nets}.
While more elaborate feature extraction methods exist, they are not considered in this work.

\subsection{Contributions}

This work is the first to systematically analyze performance of loss networks for deep perceptual loss.
Additionally, the work builds on and adds to a recent systematic analysis of loss networks for deep perceptual similarity~\cite{kumar2022surprising}.

The work provides valuable insight into which pretrained architectures and feature extraction layers perform best as loss networks for specific tasks across multiple application areas.
The analysis also reveals that deep perceptual loss does not adhere to two conventions of transfer learning.
The first convention is that better ImageNet accuracy implies better downstream performance and the second is that feature extraction from the later layers provides better performance.
Based on the findings and analysis, suggestions for future work are made including further analysis of loss networks, overarching studies for deep perceptual loss, and reevaluation of transfer learning conventions.

\section{Background}
\label{toc:literature_review}
The terminology of the subfield of perceptual loss is only sometimes used consistently between pieces of literature.
For clarity, this section is prefaced with a description of how some standard terms are used in this work.
Additionally, these terms are introduced with greater detail throughout the literature review.

\begin{itemize}
    \item A perceptual loss is a loss function that is meant to estimate a factor of human perception, often to have model predictions be closer to those of humans.
    \item A loss network is a neural network used as part of the loss calculation for training another machine learning model.
    For cohesion, this work also uses the term loss network to refer to networks used to compute image similarity.
    \item Deep perceptual loss is when the deep features of a loss network are used to calculate the perceptual loss of another machine learning model by comparing the deep features generated by the model output to those generated by the ground truth.
\end{itemize}

\subsection{Feature Extraction}

Feature extraction in machine learning is the process of converting input data into descriptive and non-redundant features.
Before feature learning became widely adopted, most feature extraction methods relied on task-related features handcrafted by experts~\cite{farias2016automatic}.
With the advancement of deep models and autoencoders~\cite{hinton2006reducing, bengio2006greedy}, feature extraction has been automated to cater to complex invariances in the data~\cite{le2013building}.
Feature extraction using deep models has been successfully used in many computer vision applications like image classification~\cite{farias2016automatic, hayakawa2016feature, lee2017deep, mohsen2018classification}, cross-media retrieval \cite{jiang2017internet}, and reinforcement learning~\cite{ha2018recurrent}.

However, determining an architecture for feature extraction and how to pretrain the network is difficult, and the best practice often varies among fields and individual tasks.
Generally, for many computer vision tasks, the trend has been to use pretrained models based on ImageNet~\cite{deng2009imagenet} which, in general, has improved downstream performance~\cite{kumar2022surprising, kornblith2019do}.
Nevertheless, it has also been shown that for unsupervised pretraining, better pretraining performance does not always lead to better downstream performance~\cite{alberti2017pitfall}.
As such it can be difficult to tell what feature extraction methods will perform well on a given task without prior evaluation.

\subsection{Perceptual Similarity Metrics}
\label{toc:perceptual_similiarty_metrics}
Measuring the similarity of images has many important applications such as image retrieval~\cite{hatzigiorgaki2003compressed}, image quality assessment~\cite{wang2004image}, and, as is described in the next section, loss calculation.
A simple approach to creating an image similarity metric is to calculate the distance between each pair of pixels, typically using the $\ell^2$ norm.
Such metrics based on the differences of corresponding pixels are called pixel-wise metrics.


Pixel-wise metrics are widely known to be flawed measurements of image similarity, especially when estimating human perception~\cite{larsen2016autoencoding}.
For example, they do not take into account the relations between different pixels or the differing importance of various pixel regions~\cite{pihlgren2020improving}.
To mitigate the flaws of pixel-wise metrics, several metrics have been proposed that estimate human perception of visual similarity, so-called perceptual similarity.
Among these perceptual similarity metrics are some that have become widely used for computer vision applications such as the Structural Similarity Index Measure (SSIM)~\cite{wang2004image}.

With the advent of deep learning, the deep features extracted by neural networks from the input images were used to do similarity comparison for content-based image retrieval~\cite{babenko2014neural} and later applied as a perceptual similarity metric~\cite{zhang2018unreasonable}.
The method has been further improved by Ding \etal{}~\cite{ding2020image} which takes inspiration from neural style transfer~\cite{gatys2016image} and uses the network to calculate both texture and content similarity.
This use of deep neural networks to calculate perceptual similarity is called deep perceptual similarity.

Deep perceptual similarity metrics commonly make use of pretrained networks.
Training the networks specifically for perceptual similarity can give a small increase in performance~\cite{zhang2018unreasonable}, especially when used together with ensemble methods~\cite{kettunen2019robust}.
However, this improvement is marginal at best, which attests to how effective the deep features of pretrained CNNs are for perceptual similarity.

A study by Kumar~\etal{}~\cite{kumar2022surprising} analyzed how different network architectures and pretraining methods affect performance in the context of deep perceptual similarity.
The study shares similarities in methodology to the present work; however, it did not examine deep perceptual loss.
The results of the study indicates that better pretraining performance does not necessarily lead to an improved perceptual similarity metric.
In fact, after a certain point, enhanced pretraining performance was shown to be detrimental to the perceptual similarity results.
The upper bound of the perceptual similarity performance as a function of ImageNet accuracy was shown to gradually increase until a certain threshold, after which the performance decreased.  
While prior studies have also shown limited correlations between the pretraining and downstream performance~\cite{alberti2017pitfall}, the clear performance increase until a certain point followed by a clear decrease is a surprising result.
Moreover, previous works have shown that the selected model and pretraining strategy substantially impact the perceptual similarity results~\cite{kumar2022surprising}.
Thus, selecting the appropriate architecture and pretraining procedure is more important than any training step involving actual perceptual similarity data.
A recent work has expanded these findings to more human similarity datasets, including semantic similarity~\cite{muttenthaler2023human}.

\subsection{Deep Perceptual Loss}

In image models trained by comparing the output to some ground truth image, that comparison is inherently a similarity measurement.
Despite the known flaws of pixel-wise metrics, they have long been the most popular method for calculating image similarity in loss calculation of machine learning models~\cite{pihlgren2020improving, larsen2016autoencoding}.
However, in recent years improvements have been made to many applications by transitioning to perceptual similarity metrics for loss calculation.
For example, SSIM has been adapted as a perceptual loss function~\cite{zhao2017loss, snell2017learning, shi2019adaptive}.


The most popular and widespread group of perceptual losses are those based on deep perceptual similarity, so-called deep perceptual loss.
Deep perceptual loss was first used with neural style transfer~\cite{gatys2016image} where features from a pretrained version of VGG-19~\cite{simonyan2014deep} were used to estimate perceptions of style and content.
These perceptions were then used as a loss to generate images with the perceived content of one image and the perceived style of another. 
Since its introduction, deep perceptual loss has been a popular tool for image-generation tasks.
It was used in the VAE-GAN (variational autoencoder, generative adversarial network) in which the discriminator acts as a loss network to facilitate higher quality image generation~\cite{larsen2016autoencoding}.
In the VAE-GAN, the discriminator is trained alongside the generator; however, other works have used pretrained loss networks, removing the need for extra training~\cite{dosovitskiy2016generating}.



Despite the growing use and popularity of deep perceptual loss, there has been relatively little effort in investigating how to best utilize the method.
Most works pick a pretrained architecture without further investigation.
Rare works investigate a few different models~\cite{dosovitskiy2016generating} or a few different feature extraction layers~\cite{mosinska2018beyond}.
While studies exist that cover how useful different models are for general feature learning applications~\cite{kornblith2019do, yosinski2014transferable, zhuang2021comprehensive} and deep perceptual similarity~\cite{kumar2022surprising}, no such studies exist for deep perceptual loss.

\subsection{Applications of Deep Perceptual Loss}
Deep perceptual similarity metrics can be directly applied to any task where measuring the similarity of images is useful such as image quality assessment and content-based image retrieval.
In the former the similarity of a distorted image to its ideal counterpart can be used as a measurement of image quality~\cite{ding2020image} and in the latter the similarity metric can be used to identify images in the database with similar content~\cite{babenko2014neural}.
Deep perceptual similarity metrics can also be directly applied to perceptual similarity datasets to evaluate how well they align with human perception of similarity~\cite{zhang2018unreasonable, ding2020image}.
Finally, deep perceptual similarity metrics can also be applied as part of a deep perceptual loss to train machine learning models for other applications.

This work studies loss networks as directly applied to image similarity as well as training with deep perceptual loss.
While deep perceptual similarity metrics have been systematically studied previously~\cite{kumar2022surprising}, this work expands the study by examining additional architectures.
Additionally, analyzing how loss network performance on perceptual similarity correlates with the performance of models trained with that loss network is interesting.
Deep perceptual loss, on the other hand, has not been systematically studied previously.
This work evaluates loss networks for the three main application areas of deep perceptual loss; image synthesis and transformation, pixel-level prediction, and image embedding.

Deep perceptual loss can be applied to any task where the output of the model being trained can be used as the input to an image network.
The most straightforward such application area is image synthesis and transformation where the model is trained to output images.
Deep perceptual loss is frequently used in this application area for tasks such as image fusion~\cite{xu2020infrared}, style-transfer~\cite{johnson2016perceptual, gatys2016image}, and super-resolution~\cite{yang2018low, ledig2017photo}.

Deep perceptual loss has also been used to improve the training of models with outputs that make predictions at the pixel level.
As these outputs share the structure of an image, deep perceptual loss can be used to compare the image-like predictions to the ground truth.
Tasks in this application area include image segmentation~\cite{mosinska2018beyond}, object detection~\cite{li2017perceptual}, and image depth prediction~\cite{liu2021perceptual}.

Not all models that are trained to output images are used for that purpose.
Instead, the deep features of the models can be used as a starting point or input for other downstream tasks.
This application area is image embedding and here deep perceptual loss may be used like in any of the two previous application areas for training a model.
However, the model is evaluated by the performance on the downstream task that makes use of the embeddings.
A prominent example of image embedding where the output image is only used for training is dimensionality reduction via autoencoding, which has been improved with the use of deep perceptual loss~\cite{pihlgren2020improving}.

For each of the application areas, this work uses a specific application as a benchmark.
Super-resolution transformation, image segmentation, and image autoencoding are used respectively as the applications for image synthesis and transformation, pixel-level prediction, and image embedding.
The chosen applications of deep perceptual loss are described in more detail below.


\subsubsection{Super-resolution}
Super-resolution is the task of transforming a low-resolution image into a high-resolution image~\cite{wang2021deep}.
The practicality of super-resolution models has been demonstrated in real-world applications such as security~\cite{rasti2016convolutional} and medical image processing~\cite{greenspan2009superresolution, saji2015superresolution}.
Deep perceptual loss has been widely adopted to train super-resolution models~\cite{johnson2016perceptual, yang2018low, rad2019targeted}.

\subsubsection{Image Segmentation}
Image segmentation involves dividing an image into different segments representing classes or instances at the pixel level.
Each pixel in the image is assigned a class or instance, and the resulting segmentation model output is a 2D lattice that can be interpreted as an image.
Deep perceptual loss has been used to train segmentation models in applications such as delineating roads in aerial images, locating cracks in roads, identifying the edges of cells in microscope images~\cite{mosinska2018beyond}, and segmenting medical images~\cite{chai2020perceptual}.

\subsubsection{Autoencoding}
Autoencoders have been used for decades as a tool for dimensionality reduction and feature learning~\cite{rumelhart1985learning, ballard1987modular}.
They have also been used for purposes beyond dimensionality reduction, such as image generation~\cite{kingma2013auto}.
Deep perceptual loss has been used to train autoencoders for image generation~\cite{larsen2016autoencoding} as well as for their original dimensionality reduction purposes~\cite{pihlgren2020improving}.

\subsection{Pretrained Torchvision Networks}
The Torchvision package~\cite{marcel2010torchvision} provides many different renowned CNN architectures that have been pretrained on the ImageNet dataset~\cite{deng2009imagenet}.
Many of these CNN architectures have won challenges and awards like the ILSVRC-challenges in 2012~\cite{krizhevsky2012imagenet} and 2014~\cite{szegedy2015going}, and the CVPR award 2017~\cite{huang2017densely}.
This section briefly covers the accomplishments, innovations, and general design of the architectures.
However, for exact implementation details, the reader is referred to the works that introduced those architectures.
The ImageNet accuracy as well as some attributes of the pretrained models used can be found in Table~\ref{tab:cnn_attributes}.
The investigated Torchvision architectures are described below.

\begin{table*}[t]
    \caption{
        Attributes of the ImageNet pretrained Torchvision~\cite{marcel2010torchvision} models used in this work.
        Shown are their accuracy on ImageNet, depth, MFLOPS for the forward pass (for a $224\times224$ pixel image), and whether the architecture has skip-connection, branches, $1\times1$ convolutions, or batch normalization.
    }
    \label{tab:cnn_attributes}
    \centering
    \small
    \begin{tabular}{l l c c c c c c c}
    \toprule
         Family & Architecture & \makecell{ImageNet\\Acc. (\%)} & Depth & MFLOPS & \makecell{Skip-\\conn.} & Branch & \makecell{$1\times1$\\conv.} & \makecell{Batch\\norm}\\\hline
        \multirow{4}{*}{\textit{VGG}}
        & VGG-11~\cite{simonyan2015very} & 69.020 & 11 & 7637 &  &  &  & \\
        & VGG-16~\cite{simonyan2015very} & 71.592 & 16 & 15517 &  &  &  & \\
        & VGG-16\_bn~\cite{simonyan2015very} & 73.360 & 16 & 15544 &  &  &  & $\checkmark$\\
        & VGG-19~\cite{simonyan2015very} & 72.376 & 19 & 19682 &  &  &  & \\\hdashline
        \multirow{3}{*}{\textit{ResNet}}
        & ResNet-18~\cite{he2016deep} & 69.758 & 18 & 1827 & $\checkmark$ &  &  & $\checkmark$\\
        & ResNet-50~\cite{he2016deep} & 76.130 & 50 & 4143 & $\checkmark$ &  & $\checkmark$ & $\checkmark$\\
        & ResNeXt-50 32x4d~\cite{xie2017aggregated} & 77.618 & 50 & 4298 & $\checkmark$ & $\checkmark$ & $\checkmark$ & $\checkmark$\\\hdashline
        \multirow{2}{*}{\textit{Inception}}
        & GoogLeNet~\cite{szegedy2015going} & 69.778 & 22 & 1516 &  & $\checkmark$ & $\checkmark$ & $\checkmark$\\
        & InceptionNet v3~\cite{szegedy2016rethinking} & 77.294 & 49 & 2850 &  & $\checkmark$ & $\checkmark$ & $\checkmark$\\\hdashline
        \multirow{2}{*}{\textit{EfficientNet}}
        & EfficientNet\_B0~\cite{tan2019efficientnet} & 77.692 & 81 & 407 & $\checkmark$ &  &  & $\checkmark$\\
        & EfficientNet\_B7~\cite{tan2019efficientnet} & 78.642 & 271 & 5308 & $\checkmark$ &  &  & $\checkmark$\\\hdashline
        \multirow{3}{*}{\textit{Other}}
        & AlexNet~\cite{krizhevsky2014one} & 56.522 & 8 & 717 &  & \\
        & DenseNet-121~\cite{huang2017densely} & 74.434 & 121 & 2899 & $\checkmark$ &  & $\checkmark$ & $\checkmark$\\
        & SqueezeNet 1.1~\cite{iandola2016squeezenet} & 58.178 & 18 & 360 &  & $\checkmark$ & $\checkmark$ & \\
    \bottomrule
    \end{tabular}
\end{table*}


\subsubsection{VGG} 

 VGG is an innovative object-detection deep model proposed by Simonyan and Zisserman in 2012~\cite{simonyan2015very}.
 It is a CNN that is one of the best object-detection models even today.
 VGG uses small-size convolution filters that allow the integration of more weighted layers.
 This work explores some of its versions, namely, VGG-11, VGG-16, VGG-19, and a version of VGG-16 that uses batch normalization (batch norm) called VGG-16\_bn.

\subsubsection{Residual Networks} 

To solve the problem of vanishing or exploding gradients in very deep networks, residual networks (ResNet) using skip connections were introduced~\cite{he2016deep}.
Skip connections connect later layers to earlier ones skipping some layers in between.
Many updated versions of ResNet were introduced later to improve the performance.
One of the subsequent versions, called ResNeXt-50, redesigned the fundamental building block of ResNet to use a multi-branch setup similar to inception networks~\cite{xie2017aggregated}.
This work uses three residual networks; ResNet-18, ResNet-50, and ResNeXt-50 32x4d.

\subsubsection{Inception Networks} 
Choosing the appropriate filter size for different layers in a CNN architecture can be challenging.
If the filter size is large, it may lose the distribution locally; if the filter is small, it may lose the information distributed globally.
Inception networks~\cite{szegedy2015going} were introduced to handle this issue by using multiple filter sizes at each layer, leading the networks to be "wider" rather than "deeper".
Inception networks have achieved state-of-the-art performance in image classification.
This work explores the first version of Inception Network developed by the Google team, \ie, GoogLeNet~\cite{szegedy2015going} and Inception Network version 3~\cite{szegedy2016rethinking}.

\subsubsection{EfficientNet} 
EfficientNet~\cite{tan2019efficientnet} is a CNN-based architecture that uses compound coefficients to scale the architecture's dimensions uniformly.
Unlike traditional methods that arbitrarily scale these factors, a set of fixed scaling coefficients is used to scale the networks in a principled way uniformly.
This method combined with neural architecture search~\cite{zoph2017neural} produced the state-of-the-art EfficientNet architectures.
In this work, the two versions of EfficientNet with the fewest and most layers are explored, \ie, EfficientNet\_B0 and EfficientNet\_B7.

\subsubsection{AlexNet} 
AlexNet is the name given to the network introduced by~\citet{krizhevsky2012imagenet} and further built upon in~\cite{krizhevsky2014one}.
While AlexNet has lower performance and robustness than more recent CNN architectures~\cite{de2021impact}, its features have proven competitive to those same networks when used as a perceptual similarity metric~\cite{zhang2018unreasonable}.

\subsubsection{DenseNet} 
DenseNet~\cite{huang2017densely} focused on using fewer parameters with densely connected layers to achieve higher accuracy.
It is composed of dense blocks.
Each layer in the dense block takes the feature maps of preceding layers as input.
This concept of DenseNet reduces the vanishing gradient problem, improves feature propagation, stimulates feature reuse, and considerably reduces the number of parameters.
In this work, the DenseNet-121 implementation is examined.

\subsubsection{SqueezeNet} 

While much of the contemporary research on deep CNNs focused on improving the performance, Iandola \etal{}~\cite{iandola2016squeezenet} focused on reducing the size of deep models by proposing SqueezeNet.
It has a similar performance to AlexNet with 50 times fewer parameters.
To achieve these benefits, SqueezeNet has reduced the kernels with size 3 to $1\times1$ convolutions, used squeeze layers to decrease the number of input channels to the kernels and downsample later in the architecture to have large activation maps with the assumption that it leads to higher classification accuracy.

\section{Loss Networks and Feature Extraction}
\label{sec:loss_nets}

This work primarily explores the effects of (1) different pretrained architectures and (2) feature extraction layers on performance.
14 architectures pretrained from ImageNet from the Torchvision framework~\cite{marcel2010torchvision} are used in this work.
Four extraction layers have been chosen for each architecture such that they represent layers that are early, semi-early, middle, and late in the convolutional layers.
Each architecture along with the feature extraction layers is detailed in Table~\ref{tab:networks_layers}.
The loss networks used for all the benchmark experiments are created by selecting a pretrained model and one of the extraction layers, resulting in 56 different loss networks.
Feature extraction from the loss networks is simple and consists of propagating activations forwards through the network to the extraction layer after which they are applied without any additional computation as defined in each benchmark.

The feature extraction layers have been limited to the convolutional layers of the networks as these are typically used for deep perceptual loss.
The convolutional layers also benefit from having no upper limit to the size of the input images, while using later layers typically requires the use of an exact input size.
Additionally, for architectures of similar design, the extraction layers are chosen such that they represent the "same" layers in each network.
For example, in the residual networks, the extraction layers are after the first ReLU as well as after the same block stacks with the difference that each residual network architecture has a varying number of layers in the different block stacks.

\begin{table}[t]
    \caption{Loss Network Architectures and Feature Extraction Layers}
    \label{tab:networks_layers}
    \begin{tabular}{l l}
    \toprule
        Architecture & Feature Extraction Layers\\\hline
        \multicolumn{2}{c}{\textit{VGG Networks}}\\
        VGG-11~\cite{simonyan2015very} & 1st, 2nd, 4th, and 8th ReLU\\
        VGG-16~\cite{simonyan2015very} & 2nd, 4th, 7th, and 13th ReLU\\
        VGG-16\_bn~\cite{simonyan2015very} & 2nd, 4th, 7th, and 13th ReLU\\
        VGG-19~\cite{simonyan2015very} & 2nd, 4th, 8th, and 16th ReLU\\
        \multicolumn{2}{c}{\textit{Residual Networks}}\\
        ResNet-18~\cite{he2016deep} & 1st ReLU, 1st, 2nd, and 4th Block Stack\\
        ResNet-50~\cite{he2016deep} & 1st ReLU, 1st, 2nd, and 4th Block Stack\\
        ResNeXt-50 32x4d~\cite{xie2017aggregated} & 1st ReLU, 1st, 2nd, and 4th Block Stack\\
        \multicolumn{2}{c}{\textit{Inception Networks}}\\
        GoogLeNet~\cite{szegedy2015going} & 1st BN, 1st, 3rd, and 9th Inception Module\\
        InceptionNet v3~\cite{szegedy2016rethinking} & 3rd BN, 1st, 3rd, and 8th Inception Module\\
        \multicolumn{2}{c}{\textit{EfficientNet}}\\
        EfficientNet\_B0~\cite{tan2019efficientnet} & 1st SiLU, 1st, 4th, and 7th MBConv\\
        EfficientNet\_B7~\cite{tan2019efficientnet} & 1st SiLU, 1st, 4th, and 7th MBConv\\
        \multicolumn{2}{c}{\textit{Other Networks}}\\ 
        AlexNet~\cite{krizhevsky2014one} & 1st, 2nd, 3rd, and 5th ReLU\\
        DenseNet-121~\cite{huang2017densely} & 1st ReLU, 1st, 2nd, and 4th Dense Block\\
        SqueezeNet 1.1~\cite{iandola2016squeezenet} & 1st ReLU, 1st, 4th, and 8th Fire Module\\
    \bottomrule
    \end{tabular}
\end{table}


\section{Benchmarks}
\label{sec:benchmarks}
In order to evaluate the effects of loss networks with different pretrained architectures and feature extraction layers this work uses one benchmark each from the application areas of image similarity, image synthesis and transformation, pixel-level prediction, and image embedding.
The four benchmarks cover the most common applications of loss networks in each area; perceptual similarity (Benchmark~1), super-resolution transformation (Benchmark~2), image segmentation (Benchmark~3), and autoencoding (Benchmark~4).
The four benchmarks are based on the experiments from \textit{The Unreasonable Effectiveness of Deep Features as a Perceptual Metric}~\cite{zhang2018unreasonable} (Benchmark~1), \textit{Perceptual Losses for Real-Time Style Transfer and Super-Resolution}~\cite{johnson2016perceptual} (Benchmark~2), \textit{Beyond the Pixel-Wise Loss for Topology-Aware Delineation}~\cite{mosinska2018beyond} (Benchmark~3), and \textit{Pretraining Image Encoders without Reconstruction via Feature Prediction Loss}~\cite{pihlgren2021pretraining} (Benchmark~4).
The benchmark experiments are run with each of the loss networks and the performance scores of those experiments are collected.
These scores are used to analyze how different attributes of the loss networks affect the downstream performance across different tasks.

The specific experiments selected for the benchmarks were based on three criteria; (i) diversity of applications, (ii) ease of implementation, and (iii) popularity.
All four works had preexisting implementations and were decided to be easy to adapt for the evaluation in this work.
As some of the authors had previously worked with Benchmark~4, this implementation was already well understood and adapted for these experiments.
Benchmarks~1 and~2 both have had a significant impact with citations in the thousands.
While Benchmarks~3 and~4 do not boast as many citations they have been referenced plenty, especially considering that they represent less popular application areas than the other two.

Each benchmark experiment along with the datasets and performance scores they use are detailed in the subsections below.
For complete technical details on the benchmarks, it is recommended to read the original works.
Some changes have been made to the benchmarks from the original experiments, which are also detailed in the subsections.

\subsection{Benchmark 1: Perceptual Similarity}
One popular benchmark of perceptual similarity is the Berkeley-Adobe Perceptual Patch Similarity (BAPPS) dataset~\cite{zhang2018unreasonable}.
BAPPS contains human judgments of image similarity on a host of distorted image patches and is used to evaluate similarity metrics by measuring how closely they align with these judgments.
BAPPS was introduced in a paper called \textit{The Unreasonable Effectiveness of Deep Features as a Perceptual Metric}~\cite{zhang2018unreasonable} which used the dataset to compare a few deep perceptual similarity metrics trained by different means against a variety of traditional metrics.
The paper shows that ImageNet pretrained deep perceptual similarity metrics outperform unsupervised pretraining~\cite{krahenbuhl2016data}, semi-supervised pretraining~\cite{pathak2017learning, zhang2017split, noroozi2016unsupervised, donahue2017adversarial}, and rule-based methods on perceptual similarity on BAPPS and come close to human performance without any task-specific training.


The BAPPS dataset consists of $197000$ $64\times64$ image patches from several sources~\cite{deng2009imagenet, scharstein2001taxonomy, bychkovsky2011learning, dang2015raise, agustsson2017ntire, su2017deep} along with versions of those patches that have been distorted by common image augmentations and algorithms such as autoencoding, super-resolution upscaling, frame interpolation, deblurring, and colorization.
The BAPPS dataset is split into two parts; Two Alternative Forced Choice (2AFC) and Just Noticable Differences (JND).
The former consists of triplets with an image patch and two distorted versions of that image patch.
Each 2AFC triplet is labeled by the fraction of human judges that considered each distorted version to be more similar to the original.
The JND part consists of image pairs that are labeled by the fraction of human judges that considered the pair to be the same image after a brief viewing.
During judgment collection, there were additional image pairs that were actually the same as well as completely different to make the task non-trivial.
The 2AFC part is further split into training and test sets while JND is purely used for testing.
For Benchmark~1 no training is performed so only the test sets are used.

Benchmark~1 consists of evaluating loss networks on the BAPPS test sets according to the procedure set out in the paper where it was introduced.
This follows the same general procedure as most deep perceptual similarity evaluations, except more architectures and extraction layers are considered.
The deep features extracted from each loss network are used to calculate a similarity between two images $x$ and $x_0$ according to 
\begin{equation}
    \label{eq:benchmark1_distance}
    d(x,x_0) = \sum^N_n\frac{1}{H_nW_n} w_n\odot||z_n-z_{0n}||_2^2
\end{equation}
where $z$ and $z_0$ are the channel-wise unit-normalized feature extractions from the loss network and $w$ are the importance weights.

The performance scores used for Benchmark~1 are the same as those used for the BAPPS dataset; the average 2AFC score and mAP\% on the JND part.
The 2AFC score for a single triplet is described as
\begin{equation}
    \label{eq:2afc score}
    \text{2AFC}(x, x_0, x_1, J) = 
    \begin{cases}
        J,& \text{if } d(x, x_1) < d(x, x_0)\\
        1-J,& \text{otherwise}
    \end{cases}
\end{equation}
where $d$ is a similarity metric, $x$, $x_0$ and $x_1$, an image and its distorted versions, and  $J$ the fraction of judgments that consider $x_1$ more similar to $x$ than $x_0$.

\subsection{Benchmark 2: Super-resolution}
The most common application of deep perceptual loss is super-resolution.
One successful work that shows how deep perceptual loss can improve the performance of super-resolution applications is \textit{Perceptual Losses for Real-Time Style Transfer and Super-Resolution}~\cite{johnson2016perceptual}.
This is shown through peak signal-to-noise ratio (PSNR) and SSIM~\cite{wang2004image} measurements as well as evaluation by human subjects.
Interestingly, deep perceptual loss was outperformed by a state-of-the-art model called SRCNN~\cite{dong2015image} on PSNR and SSIM but won with a large margin on the human judgments.

The improvement is achieved by implementing an image transformation network consisting of five modified~\cite{gross2016training} residual blocks~\cite{he2016deep} and training that network with deep perceptual loss.
The performance of the image transformation network is then compared to one trained with pixel-wise loss.
The deep perceptual loss is calculated according to
\begin{equation}
    \label{eq:benchmark2_loss}
    \hspace{0pt}L_{feat}^{\phi,n} (\hat{x},x) = \frac{1}{C_nH_nW_n}||\phi_{n}(\hat{x})-\phi_{n}(x)||_2^2
\end{equation}
where $x$ is the ground truth image, $\hat{x}$ the output of the transformation network, and $\phi_n(x)$ are the features at layer $n$ of the loss network $\phi$ with input $x$.

The image transformation networks are trained on $288\times288$ image patches from the MS-COCO training set~\cite{Lin2014MicrosoftCC}, a dataset of $300000$ natural images used for object detection, dense pose estimation, keypoint detection, image captioning, and semantic, instance, and panoptic segmentation.
The networks are tested on the images in the Set5~\cite{Bevilacqua2012LowComplexitySS}, Set14~\cite{Zeyde2010OnSI} and BSD100~\cite{Huang2015SingleIS} datasets.
The three test sets are collections of high-quality images that have become standard for evaluation of super-resolution models~\cite{agustsson2017ntire}.
Since the transformation networks are fully convolutional, they are applied to the full images of the test sets.
The transformation networks are trained and tested for $\times4$ and $\times8$ upscaling.
The input images are downscaled by applying Gaussian blur with $\sigma=1.0$ and downsampling by the given factor with bicubic interpolation.


Benchmark~2 consists of the training of image transformation networks for $\times4$ and $\times8$ upscaling following the same procedure as the paper.
Training is performed using $10K$ images randomly chosen from the MS-COCO 2014 training set using a specific seed for reproducibility purposes.
The original paper performed training for $200K$ iterations, but it is unclear whether a validation step was performed and which model was finally used for inference.
In the preliminary experiments, several architectures collapsed in the final epochs and produced black images.
To avoid having a collapsed model, in the final experiments, the training has a validation step in every epoch to retain the best model according to the PSNR metric for 100 images from the MS-COCO 2014 validation set.
As commonly used in super-resolution, the performance scores for Benchmark~2 are the PSNR and SSIM scores for the Set~5, Set~14, and BSD100 datasets, with $\times4$ and $\times8$ upscaling.
Since the style-transfer applications in the original work do not have performance scores they have not been included in the benchmark.

\subsection{Benchmark 3: Image Segmentation}
The paper \textit{Beyond the Pixel-Wise Loss for Topology-Aware Delineation}~\cite{mosinska2018beyond} applies deep perceptual loss to the task of delineating curvilinear structures through semantic segmentation.
The paper shows that the addition of deep perceptual loss improves performance compared to solely using pixel-wise loss.
This is done by training the U-net architecture~\cite{ronneberger2015u} using a combination of pixel-wise binary cross entropy loss ($L_{bce}$) and deep perceptual loss ($L_{top}$).
The loss is defined as 
\begin{equation}
    \label{eq:benchmark3_loss}
    \hspace{0pt}L(x,y) = L_{bce}(x,y) + \mu L_{top}(x,y)
\end{equation}
where $x$ and $y$ are the input image and ground truth segmentation map and $\mu$ is a scalar that is set to keep the two losses at the same order of magnitude.
The deep perceptual loss is calculated as
\begin{equation}
    \label{eq:benchmark3_dpl}
    \hspace{0pt}L_{top} (x,y) = \sum_{n=1}^{N} \frac{1}{C_nH_nW_n}\sum_{m=1}^{C_n}||\phi_{n}^m(y)-\phi_{n}^m(U(x))||_2^2
\end{equation}
where $U$ is the U-net model being trained, and $\phi_n^m$ is $m$-th feature map in layer $n$ of the loss network $\phi$ with extraction layers $n\in N$ with $C_n$ channels of size $W_n \times H_n$.


Both during training and testing the U-net is applied iteratively three times to improve the prediction.
To achieve this the network is given both the input image and the output of the previous iteration as input.
The data from the previous iteration is left blank for the first iteration.

In the original work, the method is evaluated on three datasets: Cracks~\cite{zou2012cracktree}, The Massachusetts Roads Dataset (MRD)~\cite{mnih2013machine}, and the ISBI’12 challenge~\cite{arganda2015crowdsourcing}.
The evaluation scores used are completeness, correctness, and quality of delineation on the test sets as defined in~\cite{wiedemann1998empirical}.
Completeness quantifies the fraction of ground truth objects that have been correctly delineated by the model and correctness measures the fraction of predicted objects that have a matching ground truth object.
The quality metric is a balance between the two by giving the fraction of all ground truth and predicted objects that were correctly delineated.
In short, completeness, correctness, and quality are analogs to recall, precision, and critical success index respectively.

Benchmark~3 only considers evaluation on MRD with input patches resized to $224\times224$ for computational efficiency.
MRD is a set of $1171$ aerial images with corresponding segmentation maps of the roads.
The benchmark follows the training and testing procedure of the original paper.
The benchmark uses $100$ training epochs and $0.01$ as the value of $\mu$ as both of these are ambiguous in the original paper.
The performance scores for Benchmark~3 are the completeness, correctness, and quality of the trained U-nets on the MRD test set.

\subsection{Benchmark 4: Autoencoding}
In \textit{Pretraining Image Encoders without Reconstruction via Feature Prediction Loss}~\cite{pihlgren2021pretraining} deep perceptual loss is applied to autoencoder training for downstream prediction.
The work shows that image autoencoders trained with deep perceptual loss learn more useful features for downstream prediction, than those trained with pixel-wise loss.
This is achieved by adapting a previous image autoencoder setup~\cite{ha2018recurrent}, training it to encode and reproduce images, and training small Multi-Layer Perceptrons (MLP) on downstream tasks using the autoencoder embeddings as input.
The performance of the downstream MLP reflects how useful the learned embeddings are for predictions and, therefore, which loss is best to use for training the autoencoders for downstream prediction.

The loss calculation for autoencoder training is defined as
\begin{equation}
    L_{dp}(x) = \sum^m_{k=1} ||\phi(x)_k - \phi(A(x))_k||^2_2
    \label{eq:benchmark4_loss}
\end{equation}
where $x$ is the input image, $A(x)$ is the autoencoder reconstruction, $\phi$ is a loss network with $m$ extracted features.

The downstream tasks that the autoencoders are evaluated on are classification on the SVHN~\cite{netzer2011reading} and STL-10~\cite{coates2011analysis} datasets, as well as object positioning on images gathered from the LunarLander-v2 environment of the OpenAI Gym~\cite{brockman2016openai}.
For all three evaluations, the autoencoders are trained on an auxiliary set of images in the data, the MLPs are trained and validated on an 80/20 split of the training sets, and the MLPs for each autoencoder with the best validation score are tested on the test sets.
The auxiliary images are those in the extra set from SVHN and the unlabeled images from STL-10.

For Benchmark~4 evaluation is only performed on the easily accessible and often used SVHN and STL-10 datasets.
SVHN consists of $32\times32$ pixel images of house number digits that are labeled by the digit, which are scaled up to $64\times64$ images through reflection padding.
STL-10 consists of $96\times96$ pixel images of animals and vehicles taken from ImageNet, where the training and test sets consist of 4 vehicle and 6 animal classes.
The autoencoders have an embedding size of 64 and are trained for 10 epochs using the loss in Eq.~\ref{eq:benchmark4_loss} with the different loss networks.
For each autoencoder, seven different MLPs with zero, one, and two hidden layers and various layer sizes are trained for 100 epochs each.
The performance scores for the benchmark are the test set accuracies of the MLP with the best validation accuracy for each autoencoder on SVHN and STL-10.
The training epochs are reduced compared to the original work to save computation and it has been confirmed through smaller-scale experiments that this only minimally affects performance.

\section{Results and Analysis}
\label{toc:results}


The performance scores on the four benchmarks have been analyzed together with a host of different attributes to attempt to provide insights into how to decide which architecture to choose and what layers to extract features from.
The attributes that have been analyzed can be split into those that depend only on the architecture and those that depend on the architecture and extraction layer.
The attributes that depend on the architecture are, ImageNet accuracy, number of parameters, depth\footnote{Depth is measured as the maximum number of non-linearly separated matrix multiplications up to the given layer (\ie if the architecture branches and then rejoins only the longest branch is counted).}, flops in the forward pass, and whether the architecture uses skip-connections, branching, $1\times1$-convolutions, or batch norm.
The attributes that also depend on the extraction layer are the number of parameters and depth before the extraction layer as well as the number of features and channels at the extraction layer.
Additionally, some attributes derived from these were used, such as the fraction of the architecture's total parameters that exist up until the extraction layers.
Finally, the performance scores of the four benchmarks were compared with each other to identify potential task-independent trends.


From the data gathered three primary findings related to how to select a loss network were identified.
The three findings relate to what impact architecture, extraction layer, and pretraining accuracy have on the performance scores.
Those three findings are summarized below and then expanded on in their respective subsections.
\begin{enumerate}
    \item In general, the VGG networks without batch norm and SqueezeNet perform well for most tasks if the correct layers are used.
    \item Using the correct extraction layer is at least as important as selecting the architecture.
    \item There is no simple correlation between an architecture's performance on ImageNet and its performance as a loss network.
\end{enumerate}
The three findings also show that some correlations, which are typically assumed in the field of transfer learning, do not apply to deep perceptual loss and similarity.
The analysis of the implications of these results is further expanded in the Discussion section.
Information on how to access the raw data, including a spreadsheet that can be used to create figures similar to Figs.~\ref{fig:performance_in_order}, \ref{fig:svhn+stl-10_vs_extraction-point}, and~\ref{fig:2afc_vs_parameters}, can be found in the supplementary material as detailed in the Appendix.

\subsection{Impact of Architecture}
To gain insight into what architecture attributes are beneficial for a loss network, the performance of all the architectures is analyzed.
In Table~\ref{tab:architecture_ranking}, a summary of the performance of all architectures is presented.
For each performance score, the architectures are ranked according to the performance of the best extraction layer between 1 (best) and 14 (worst).
The table presents the average ranking of each architecture per benchmark as well as the overall average (each benchmark is weighted equally).

\begin{table}[t]
    \caption{Rankings of the best loss network for each architecture on the performance scores averaged per benchmark}
    \label{tab:architecture_ranking}
    \begin{center}
    \begin{tabular}{l r r r r r}
    \toprule
        & \multicolumn{5}{c}{Average ranking per benchmark} \\
        Architecture & 1 & 2 & 3 & 4 & All\\\hline
        \multicolumn{1}{l}{\textit{VGG Networks}}\\
        VGG-11~\cite{simonyan2015very} & \textbf{2.5} & 2.33 & 2.67 & 6 & \textbf{3.38}\\
        VGG-16~\cite{simonyan2015very} & 4.5 & 6.75 & 3.67 & \textbf{1.5} & 4.10\\
        VGG-16\_bn~\cite{simonyan2015very} & 4.5 & 11.08 & 11.33 & 9.5 & 9.10\\
        VGG-19~\cite{simonyan2015very} & 7 & 4.25 & \textbf{1.67} & \textbf{1.5} & 3.60\\
        \multicolumn{1}{l}{\textit{Residual Networks}}\\
        ResNet-18~\cite{he2016deep} & 5.5 & 7.08 & 12.67 & 8 & 8.31\\
        ResNet-50~\cite{he2016deep} & 10 & 7.33 & 6.67 & 8.5 & 8.13\\
        ResNeXt-50 32x4d~\cite{xie2017aggregated} & 11.5 & 5.67 & 5.67 & 9.5 & 8.08\\
        \multicolumn{1}{l}{\textit{Inception Networks}}\\
        GoogLeNet~\cite{szegedy2015going} & 8.5 & 9.67 & 10.33 & 5.5 & 8.50\\
        InceptionNet v3~\cite{szegedy2016rethinking} & 10.5 & 11.83 & 4.67 & 11 & 9.50\\
        \multicolumn{1}{l}{\textit{EfficientNet}}\\
        EfficientNet\_B0~\cite{tan2019efficientnet} & 9 & 9.92 & 11.67 & 8.5 & 9.77\\
        EfficientNet\_B7~\cite{tan2019efficientnet} & 12.5 & 8.83 & 3 & 8.5 & 8.21\\
        \multicolumn{1}{l}{\textit{Other Networks}}\\
        AlexNet~\cite{krizhevsky2014one} & 5 & 8.58 & 9.33 & 10 & 8.23\\
        DenseNet-121~\cite{huang2017densely} & 10.5 & 10.5 & 13.67 & 8.5 & 10.79\\
        SqueezeNet 1.1~\cite{iandola2016squeezenet} & 3.5 & \textbf{1.17} & 8 & 8.5 & 5.29\\
    \bottomrule
    \end{tabular}
    \end{center}
\end{table}

Across all four benchmarks two VGG networks without batch norm place in the top three when averaging the rankings of all performance scores.
Interestingly the VGG-16 network with batch norm places in the bottom four on all benchmarks except perceptual similarity (Benchmark~1).
The only network besides the VGG networks that do not use batch norm is AlexNet, which gets an average of 8th place over all four benchmarks.
Another architecture that performs well is SqueezeNet which has an average performance on the delineation (Benchmark~3) and autoencoder training (Benchmark~4) scores but is second best at perceptual similarity (Benchmark~1) and the best at super-resolution (Benchmark~2).
SqueezeNet is also a good option in all benchmarks when looking for performance as well as low computational needs.
It is also worth noting that besides adding batch norm to VGG, the architectures within the same basic template (VGGs, ResNets, and EfficientNets) perform similarly, with little indication of which would be the better choice.

\subsection{Impact of Extraction Layer}

\begin{figure}[t] 
    \includegraphics[width=\columnwidth]{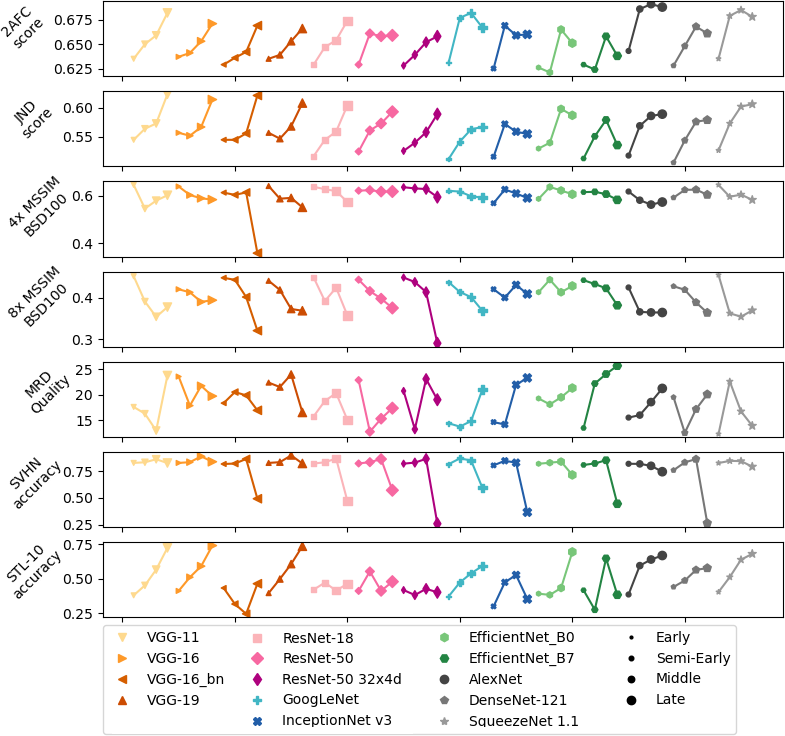}
    \caption{
        The results of each loss network ordered by extraction layer (earliest to latest) for some performance scores for each benchmark.
    }
    \label{fig:performance_in_order}
\end{figure}

The selection of where the features used for loss calculation are extracted in the network has a huge impact on the performance across all performance scores and architectures.
For all performance scores, selecting the best extraction layer of the worst architecture will give around the same performance as selecting the worst extraction layer of the best architecture.
This shows that selecting the extraction layer for the loss network is as significant as selecting which model to use.
In Table~\ref{tab:best_layers} the best layer for each architecture is shown for some performance score of each benchmark.
The architecture with the best performance for the performance score is underlined.
In Figure~\ref{fig:performance_in_order} the performance of each architecture is shown for all their extraction layers for some of the performance scores.

\begin{table*}[t]
    \caption{The feature extraction layer with the best performance of each architecture for some performance scores of each benchmark. The letters indicate extraction layers with the best loss network overall being underlined (E: Early, S: Semi-Early, M: Mid, L: Late).}
    \label{tab:best_layers}
    \begin{center}
    \begin{tabular}{l c c c c c c c}
    \toprule
        & \multicolumn{7}{c}{The best performing extraction layer per performance score.} \\
        &  \multicolumn{2}{c}{Benchmark~1} & \multicolumn{2}{c}{Benchmark~2} & \multicolumn{1}{c}{Benchmark~3} & \multicolumn{2}{c}{Benchmark~4}\\
        Architecture  & 2AFC  & JND  & \makecell{4x MSSIM\\BSDS100}  & \makecell{8x MSSIM\\BSDS100}  & MRD (All)  & SVHN  & STL-10\\\hline
        \multicolumn{1}{l}{\textit{VGG Networks}}\\
        VGG-11~\cite{simonyan2015very}  & L  & \underline{L}  & \underline{E}  & E  & L  & M  & L\\
        VGG-16~\cite{simonyan2015very}  & L  & L  & E  & E  & E  & M  & \underline{L}\\
        VGG-16\_bn~\cite{simonyan2015very}  & L  & L  & M  & E  & S  & M  & L\\
        VGG-19~\cite{simonyan2015very}  & L  & L  & E  & E  & \underline{M}  & \underline{M}  & L\\
        \multicolumn{1}{l}{\textit{Residual Networks}}\\
        ResNet-18~\cite{he2016deep}  & L  & L  & E  & E  & M  & M  & S\\
        ResNet-50~\cite{he2016deep}  & S  & L  & S  & E  & E  & M  & S\\
        ResNeXt-50 32x4d~\cite{xie2017aggregated}  & L  & L  & E  & E  & M  & M  & M\\
        \multicolumn{1}{l}{\textit{Inception Networks}}\\
        GoogLeNet~\cite{szegedy2015going}  & M  & L  & E  & E  & L  & S  & L\\
        InceptionNet v3~\cite{szegedy2016rethinking}  & S  & S  & S  & M  & L  & S  & M\\
        \multicolumn{1}{l}{\textit{EfficientNet}}\\
        EfficientNet\_B0~\cite{tan2019efficientnet}  & M  & M  & S  & S  & L  & M  & L\\
        EfficientNet\_B7~\cite{tan2019efficientnet}  & M  & M  & S  & E  & L  & M  & M\\
        \multicolumn{1}{l}{\textit{Other Networks}}\\
        AlexNet~\cite{krizhevsky2014one}  & \underline{M}  & L  & E  & E  & L  & E  & L\\
        DenseNet-121~\cite{huang2017densely}  & M  & L  & M  & E  & L  & M  & L\\
        SqueezeNet 1.1~\cite{iandola2016squeezenet}  & M  & L  & E  & \underline{E}  & L  & S  & L\\
    \bottomrule
    \end{tabular}
    \end{center}
\end{table*}

Interestingly, for most performance scores the extraction layer which performs best is similar across architectures.
This indicates that the selected extraction layers are roughly equivalent which is desired and reinforces the existing consensus that two deep networks will learn similar features at similar layers along their depth.
However, the spread of which extraction layers give the best performance on the different performance scores goes against the common practice of using the last or later layer for feature extraction.
For a majority of architectures, the earliest extraction layer is the best for the super-resolution experiments (Benchmark~2) on all performance scores, while the last extraction layer gives the highest downstream accuracy on STL-10 (Benchmark~4).

So, while there exists some agreement between architectures as to which extraction layer to use, that layer depends on the task and dataset.
While the gathered data does not give a conclusive way to predict which extraction layer will be best for a given task and dataset, there are some trends.

Selecting an extraction layer for a loss network determines what features will be compared during the calculation of loss or similarity.
When optimizing the loss, the output will trend towards an image that gives rise to similar activations at the extraction layer.
Deeper layers of networks are known to represent higher-level features, so the extraction layer directly affects which type of features will be emphasized in image generation.
Extracting from early layers means that the smaller pixel level patterns affect loss more while from later layers means that the general content and structure of the image affect the loss more.
Whether it is best to optimize images on lower or higher-level features depends on the task.
For example, in the super-resolution experiments (Benchmark~2) for all three test datasets, the optimal extraction layer was early.
Since super-resolution is a task where the individual pixels matter a lot, it is not unexpected that earlier extraction layers that are focused on low-level features are best.

\begin{figure}[t] 
    \includegraphics[width=\columnwidth]{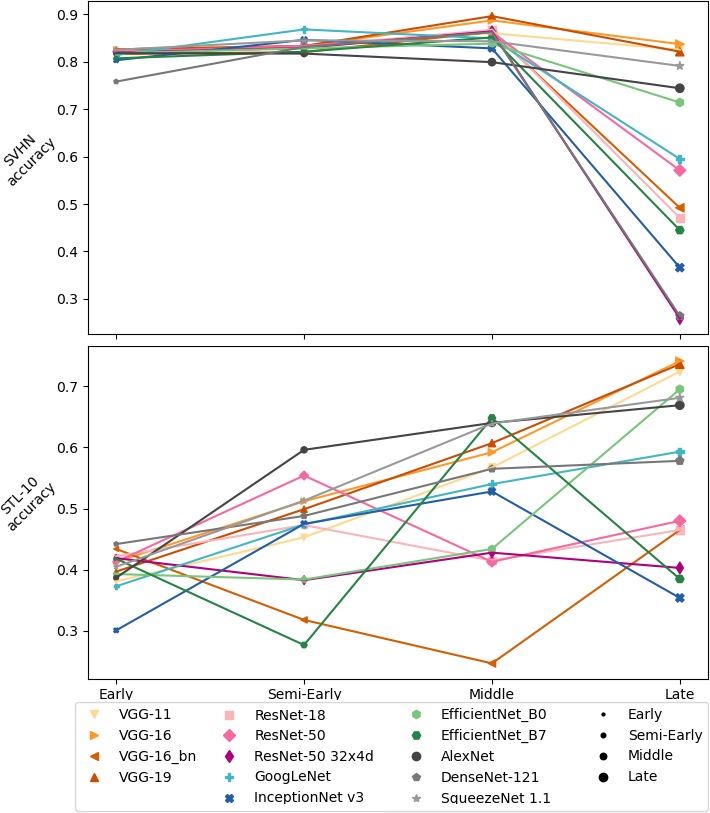}
    \caption{
        The performance of all loss networks on Benchmark~4 as measured by the downstream accuracy on SVHN and STL-10 where the loss networks have been grouped by extraction layer.
        More figures like these can be quickly generated in the supplementary spreadsheet, using any combination of investigated attributes and performance scores.
    }
    \label{fig:svhn+stl-10_vs_extraction-point}
\end{figure}

For the task of training an autoencoder for downstream predictions (Benchmark~4) later extraction layers perform better.
Likely this is due to classification tasks relying on higher-level features which means that autoencoders trained to replicate images that are similar in the later layers, would therefore also encode information relevant to higher-level features.
The difference between the two evaluated datasets, SVHN and STL-10, is also noteworthy.
For SVHN accuracy most architectures performed the best using the middle extraction layer, compared to STL-10 accuracy for which the late extraction layer was preferred. 
This difference is visualized in Fig.~\ref{fig:svhn+stl-10_vs_extraction-point} which shows the performance of the different loss networks, grouped by extraction layer, on the two datasets.
This is notable since STL-10 is derived from a subset of ImageNet. 
SVHN on the other hand is the adjusted close-up images of house number digits, which are different from the typical photos in the ImageNet dataset.
This also fits into the idea that the extraction layer should match the task since the features in the later layers are expected to be more specific to the pretraining dataset and therefore more useful for extracting features of a similar dataset.
When selecting an extraction layer it is therefore worth considering how similar the dataset is to the one used for pretraining the loss networks.

\begin{figure}[t] 
    \includegraphics[width=\columnwidth]{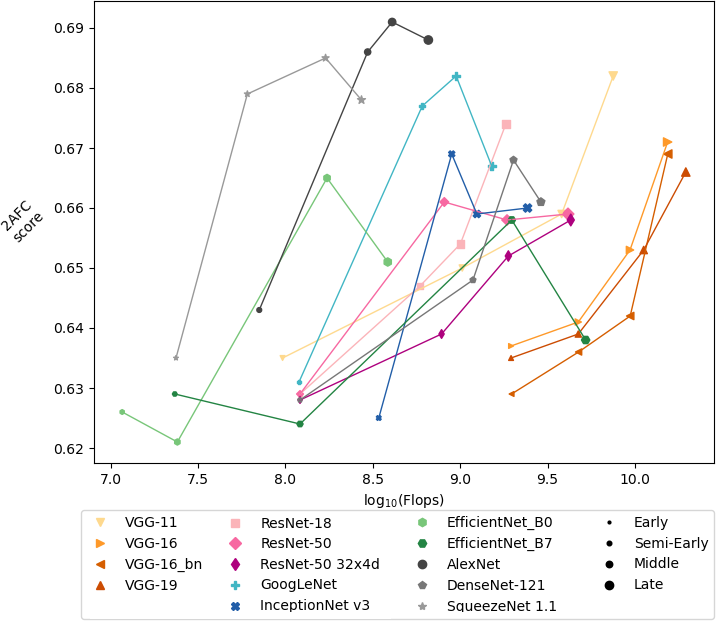}
    \caption{
        The performance of all loss networks on the 2AFC split of BAPPS compared to the $log_{10}$ amount of flops in a forward pass of that loss network.
    }
    \label{fig:2afc_vs_parameters}
\end{figure}

Another consideration when selecting an extraction layer (and architecture) is the computational demands during training.
Some of the architectures that were evaluated require high amounts of computation on top of that required for the model that is being trained.
For training smaller models this could potentially increase the computation power needed to train an order of magnitude.
Using earlier extraction layers means smaller loss networks, which reduce this additional computation.
Taking into account that the later extraction layers do not always perform better, this makes the argument for using earlier extraction layers stronger.
This is illustrated in Fig.~\ref{fig:2afc_vs_parameters} where, for each loss network, the 2AFC score is plotted against the $log_{10}$ amount flops for the forward pass.
It is clear that selecting an earlier extraction layer can potentially reduce the computation requirements by orders of magnitude.

\subsection{Loss Network Performance vs. ImageNet Accuracy}
\begin{figure}[t] 
    \includegraphics[width=\columnwidth]{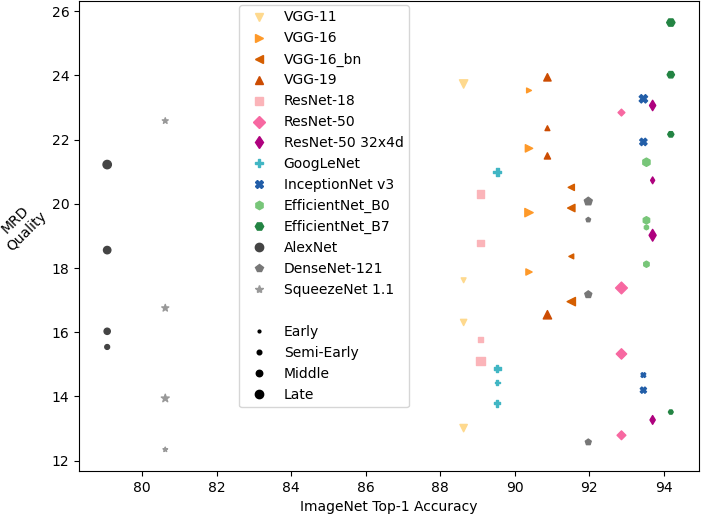}
    \caption{
        The MRD Quality performance of each loss network compared to the ImageNet top-1 accuracy or the pretrained models.
    }
    \label{fig:mrd-quality_vs_imagenet-top1-acc}
\end{figure}

For the architectures and extraction layers tested in this work, there does not seem to be any strong correlation between the ImageNet accuracy of the loss network and the downstream performance of the models trained with them.
The positive linear correlation between ImageNet accuracy and downstream performance that is often expected in computer vision transfer learning was absent in all performance scores.
This has previously been shown to hold for deep perceptual similarity~\cite{kumar2022surprising}, and now also seems to hold for deep perceptual loss.
However, the upper bound of perceptual similarity as a function of ImageNet performance that was reported in that work has not been replicated or refuted for deep perceptual loss.
It would likely require an evaluation of orders of magnitude more loss networks to do so.
The performance of the best loss network for each architecture on MRD Quality compared to the ImageNet top-1 accuracy is shown in Fig.~\ref{fig:mrd-quality_vs_imagenet-top1-acc}.

\section{Discussion}

The discussion of this work is divided into four subsections.
The first suggests an approach to implementing a suitable loss network for a given task and dataset based on the findings in this work and others.
The second discusses the two transfer learning conventions that do not hold for deep perceptual loss and what this entails for the larger field.
The third describes the limitations of this work and what weaknesses there may be in the findings.
The final subsection suggests promising directions for future research, both regarding deep perceptual loss and the wider field of transfer learning.

\subsection{Implementing a Suitable Loss Network}

This work evaluates $14$ different ImageNet pretrained architectures with four different extraction layers on four benchmarks.
The results of this evaluation, combined with results from previous works, provide insight into how to implement a suitable ImageNet pretrained loss network for a given task.

When it comes to the choice of architecture, the VGG networks without batch norm perform well on all four benchmarks.
Out of them, VGG-11 performs best on average, though both VGG-16 and~19 have benchmarks where they outperform the other VGG networks.
While it may be interpreted that batch norm leads to poor performance it cannot be stated whether this generalizes beyond the VGG networks.
Additionally, in Torchvision the models with batch norm use a different pretraining procedure than those without, specifically they use a larger learning rate.
This means that even if there was a clear difference between the architectures with and without batch norm, this difference might be due to the training procedure rather than the architecture itself.
SqueezeNet also performs well on Benchmarks~1 and~2 which focus more on small-scale features, especially super-resolution where focus on the large-scale features is not as necessary as they are part of the input to the image transformation network.
SqueezeNet has average performance on the remaining benchmarks.
Despite these clear indicators of some architectures performing better, it is difficult to generalize their findings to which attributes of these networks make them perform well.
There is no strong correlation in the gathered data between different architecture attributes and loss of network performance.
For most attributes, this lack of correlation may be due to the size of the data, and studies targeting specific attributes may find those.
However, it seems clear from the data that the total depth and number of parameters of a model do not affect its downstream performance as a loss network.

Some of the most novel findings of this work regard the choice of extraction layer for deep features of the loss networks.
The choice of feature extraction layer has at least as much impact on loss network performance as the choice of pretrained architecture.
It is also interesting to note that the extraction layer that performs best on a performance score is correlated between different architectures.
So if one architecture performs well on a task with a certain feature extraction layer, it is likely that other architectures will perform, relatively, well on the task with feature extraction at a similar relative depth.
Though, the optimal feature extraction depth does not correlate between benchmarks, nor between tasks if the tasks are substantially different.

Some inferences can be made as to which feature extraction layers will perform well on a task based on the parameters of that task.
Most performance scores on Benchmark~2 are improved for the earlier layers.
One factor is that the performance is measured with scores focusing on low-level features, which likely align closer to the features detected by earlier layers.
Another factor is that the input to the image transformation network contains the higher level structures, learning to replicate those structures should be easier as this can be achieved with the identity function, while the pixel-level differences are more difficult to learn.
In such cases, early layers perform better.
If the performance scores are less dependent on low-level structures, such as classification accuracy, or if the relevant high-level structures are not included in the input, such as image segments, later layers likely perform better.
Additionally, how similar the pretraining data is to the task data likely also affects which extraction layer is more suitable.
This can be observed in Benchmark~4, where the later extraction layers in general perform well on STL-10 which is derived from the pretraining dataset, while scores on SVHN, which is drawn from a substantially different distribution, are better for earlier extraction layers.
This is likely since the later layers have learned features that are specifically useful for the classes in ImageNet, and thus do not generalize well to SVHN.

An aspect that is not explored in this work is how the loss network is pretrained.
Kumar \etal{}~\cite{kumar2022surprising} showed that performance on the BAPPS perceptual similarity dataset could be improved by pretraining models on ImageNet with other parameters and for fewer epochs than what would give optimal ImageNet accuracy.
It is possible that such specialized pretraining could improve performance on deep perceptual loss tasks as well.
However, since deep perceptual loss tasks require training another model with the loss network, such experiments would likely be prohibitively computationally expensive.
Additional pretraining on the distribution of the downstream task might lead to more suitable deep features in the later layers as was observed in Benchmark~4 with SVHN and STL-10 accuracy.

Another interesting finding is that, besides VGG networks generally performing well, the performance of a loss network on one benchmark does not predict its performance on another.
This means that for a new task, it can be difficult to predict which loss networks should perform well and likely several will have to be tested.
If computation is limited, evaluating feature extraction layers on VGG networks is recommended.
If more computation is available it is also worth exploring some additional architectures with similar relative depth of the feature extraction layers as the best-performing VGG networks.
As pretraining architectures can be orders of magnitude more expensive than evaluating them, this is only recommended in circumstances where resources are plentiful, slight gains in performance are important, and the model will be extensively used for inference.

There are other potential methods for implementing better-performing loss networks that have not been analyzed in this work.
Some of those methods are presented later in Subsection~\ref{sub:future_work}.

\subsection{Breaking Transfer Learning Conventions}

The field of transfer learning has many conventions and accepted good practices.
For some problems and conventions, there is extensive empirical support, though the conventions are often followed beyond these cases.
The results in this work show that deep perceptual loss breaks two commonly held transfer learning conventions.
First, it extends the findings on perceptual similarity by Kumar \etal{}~\cite{kumar2022surprising} that ImageNet accuracy is not positively correlated with downstream performance, to deep perceptual loss.
Second, it demonstrates that for deep perceptual loss and similarity the later layers are not necessarily better for feature extraction.
Both of these findings are expanded upon below.

In computer vision, it is a common convention that improved ImageNet accuracy leads to better downstream performance on transfer learning tasks and this convention has been supported by trends~\cite{kumar2022surprising} and specific studies~\cite{kornblith2019do} for some problems.
However, on no benchmark can a positive correlation between the ImageNet accuracy of a model and its performance as a loss network be found.
Kumar \etal{}~\cite{kumar2022surprising} also shows that downstream performance improves with ImageNet accuracy until some threshold accuracy after which downstream performance decreases as ImageNet accuracy increases.
Though to find this correlation several thousands of loss networks were examined.
No such correlation could be found in the data gathered in this work, though likely if more loss networks were examined such patterns would reveal themselves.
Another interesting point is that ImageNet accuracy is correlated with the depth of the network, but this correlation cannot be found in the results of this work either.
This lack of correlation with depth might be part of the reason why deep perceptual loss and similarity break this transfer learning convention.
Especially, since it was also shown that the perceptual similarity scores of shallower architectures do not decay as much with increased ImageNet accuracy as their deeper counterparts~\cite{kumar2022surprising}.

In transfer learning in general it is also often held that the feature extraction should be performed in the later layers for better performance as these layers contain more advanced features~\cite{zhuang2021comprehensive}.
However, the results of this work show that for deep perceptual loss and similarity the best layer for feature extraction is heavily dependent on the task and dataset.
Since the loss networks are applied to the task of loss calucaltion without fine-tuning on that task, it is likely that the later layers contain more specialized features that are not generally applicable, leading to worse performance for some tasks.
In most other transfer learning tasks the model that uses the extracted features is trained and can therefore learn low weights for features that are not useful for the application.
Additionally, it is clear that for tasks where lower-level features are more important, training with earlier layers on which such features have greater impact is likely a good choice.

\subsection{Limitations of The Study}
The findings of this work are limited by a number of factors.
The limited number of architectures tested is likely one of the reasons why no conclusion can be drawn relating to some attributes.
The datasets used were mostly limited, especially within tasks, leading to the interpretations related to changing datasets being weak.
The selected extraction layers for each architecture are assumed to be equivalent, however, this might not be the case.
Some architectures might perform worse simply due to the features at the selected extraction layers being unsuitable for the tasks.
However, in transfer learning settings it is commonly held that slight differences in extraction layers only lead to minor changes in performance, especially if fine-tuning is used~\cite{yosinski2014transferable}.
This reasoning might extend to deep perceptual loss as well, though further study would be needed.

Another potential problem with the work is the performance scores used in Benchmark~2.
The performance scores, PSNR and SSIM, are the same as in the original work~\cite{johnson2016perceptual}.
However, these metrics have been called into question when it comes to how well they represent the human perception of similarity~\cite{zhang2018unreasonable}.
This shortcoming is also noted in the original work, where their use is justified as a way to identify differences between losses rather than an attempt at showing state-of-the-art performance.
In addition to this, it could be argued that performing well on both of these metrics might be a good indicator of quality even if it does not correlate directly with human perception of quality.
However, the preference for earlier extraction layers on Benchmark~2 is likely due to using performance scores that compare local and low-level features.
The obvious alternatives to using such metrics would be to use deep perceptual similarity or human judgments, the former of which will bias the results towards the loss network chosen for similarity and the latter of which is expensive.
Interestingly, there is no correlation between performance on the super-resolution subset of BAPPS and performance on the super-resolution metrics in Experiment~3, which seems to further indicate that the low-level metrics used do not correspond well to human perceptions of quality.

\subsection{Future Work}
\label{sub:future_work}

This work provides initial insight into how to implement suitable loss networks using pretrained architectures for a given task.
However, there are some parameters that were unexplored in this work, and of those that were explored more insight is needed.
Additionally, more exploration of deep perceptual loss and similarity is needed, including studies that aggregate knowledge regarding the method.
This work also has implications for the transfer learning field at large which suggests that further studies into other transfer learning applications are of interest.
Below, these suggestions for future work are elaborated on.

While this work has shown that the popular VGG architectures are a good choice for loss networks, more data is needed to understand what attributes in general make an architecture perform well as a loss network.
Ablation studies focusing on specific architecture attributes might provide further insight.
However such studies would likely be computationally expensive in comparison to the usefulness of their findings, as which architectures perform well are not necessarily consistent between tasks.
A more beneficial approach might be to find if performance on certain deep perceptual loss tasks is correlated with other tasks and if those tasks could be used as a proxy for evaluating loss networks.

The results of this work indicate that the optimal feature extraction layer for deep perceptual loss and similarity can be predicted based on the task and pretraining data.
Whether these indications hold on tasks and datasets beyond those evaluated in this work would require additional experiments.
As is discussed below, such evaluations may be useful in the larger transfer learning domain.

Other important aspects of pretrained loss networks that are not covered by this work are the pretraining procedure and dataset.
Kumar \etal{}~\cite{kumar2022surprising} showed that pretraining on ImageNet specifically for performance on perceptual similarity rather than accuracy can give significant improvement on perceptual similarity datasets.
However, attempting to explore the same for deep perceptual loss would likely be orders of magnitude more computationally expensive.
Additionally, it has recently been shown that the pretraining dataset and task can have a significant impact on alignment with human perception of similarity~\cite{muttenthaler2023human}.

In the field of contrastive learning, it has been shown that ImageNet transfer learning is not as beneficial in image domains that are sufficiently different from natural scenes, such as medical images~\cite{chandra2023self}.
As a supplement, pretraining on images from the application domain is being investigated~\cite{chandra2023self}.
Such supplementary pretraining could likely benefit deep perceptual loss applications as well.


Beyond using pretrained loss networks, another common approach to deep perceptual loss is to train the loss network along with the network performing the task~\cite{larsen2016autoencoding}.
This is often done in a generative adversarial network setup where the discriminator is also used as a loss network.
These methods are more difficult to evaluate and compare as they are typically designed to be task-specific and would require extensive computation for a systematic evaluation as has been performed in this work.

A challenge for many of the presented directions of future analysis of loss networks is the computational demands needed for systematic evaluation.
While the increasingly widespread use of deep perceptual loss might justify such studies, there are less demanding ways to aggregate knowledge of the method.
As of yet no extensive survey covering either improvements or applications of deep perceptual loss exists.
Such a survey would likely be highly beneficial to the field as it could identify patterns of when the method performs better.
Another strong benefit would be to unify the body of work and to spread insight between application areas that may otherwise be isolated from each other.

The findings of this work also suggest further work in the field of transfer learning.
While it has been clearly demonstrated that increased ImageNet accuracy is correlated with increased performance for some cases~\cite{kumar2022surprising, kornblith2019do}, it has also been shown that ImageNet pretraining is not as beneficial in other cases~\cite{raghu2019transfusion}.
The Pareto front between ImageNet accuracy and perceptual similarity performance found by Kumar \etal{}~\cite{kumar2022surprising}, likely exists for other tasks as well.
It is well established in multi-task learning that a solution that is optimal for one task is rarely optimal for another~\cite{ma2020efficient}.
As such Pareto fronts are likely to exist in most transfer learning settings.
Studies probing when ImageNet pretraining is beneficial and at which accuracy the optimal downstream performance is reached would be beneficial to the field at large.

Another transfer learning convention in need of reevaluation is that the later layers are best suited for feature extraction.
This work showed that this convention does not hold for deep perceptual loss and similarity.
Likely this is the case for other transfer learning methods.
Additionally, other findings relating to which layer is optimal and how the optimal layer correlates between layers might also generalize to the larger field of transfer learning.
To investigate these conventions, surveys are likely a useful tool.

\section{Conclusion}
\label{toc:conclusion}


Large-scale systematic testing and analysis of loss networks with varying architectures and extraction layers have been performed.
The results and analysis point to the three primary findings.
First, selecting the extraction layer for features is at least as important as selecting the architecture, and good practice for making this selection is suggested.
Secondly, while no general rule for selecting the architecture could be identified, the VGG networks without batch norm are a good choice.
Thirdly, there is no simple correlation between architecture attributes such as ImageNet accuracy, depth, and the number of parameters and downstream performance when used as a loss network.

The results also reinforce and expand earlier works showing that two established conventions within the field of transfer learning do not apply to deep perceptual loss and similarity.
The conventions in question are that the final layers are the best candidates for feature extraction and that better ImageNet accuracy implies better downstream transfer learning performance.
Further studies into these conventions and when they are applicable are needed.

\section*{Acknowledgment}
\noindent This work required a lot of computation to complete, which was provided through the GPU data lab of Lule\aa~University of Technology.

We would also like to thank Christian G\"unther for help with setting up the coding and execution environments, and Prof. Elisa Barney Smith for her in-depth peer-review.

\bibliography{biblio}
\bibliographystyle{IEEEtranN}

\newpage
\onecolumn
\appendix

This work has gathered and aggregated more data than can be usefully fit into the main paper.
Instead, this data is made available as supplementary material.
The supplementary material is available in three different versions: An online spreadsheet\footnote{\url{bit.ly/loss-network-analysis}}, three ancillary \textit{.csv} files published alongside this work, and this appendix.

Each version of the supplementary material serves a different purpose.
The online spreadsheet contains the raw attributes and performance scores as well as the derived attributes, aggregated results, and automatic generation of scatterplots similar to those found in this work.
This facilitates quick access to and analysis of the data for the readers.
The ancillary files contain the same raw data and derived attributes as the spreadsheet and are meant to guarantee that the raw data is available even if the online spreadsheet at some point becomes unavailable.
The appendix contains the raw performance scores of the four benchmarks so that they may be easily accessible while reading.

The performance scores for the different architectures and extraction layers for each of the four benchmarks are available in the following tables: Table~\ref{tab:benchmark1_raw_data} (Bencmark~1), Tables~\ref{tab:benchmark2_raw_data_4x_psnr}--\ref{tab:benchmark2_raw_data_8x_mssim} (Benchmark~2), Table~\ref{tab:benchmark3_raw_data} (Benchmark~3), and Table~\ref{tab:benchmark4_raw_data} (Benchmark~4).

\begin{table*}[ht]
    \caption{2AFC and JND scores of the BAPPS dataset for different loss networks used to measure deep perceptual similarity}
    \label{tab:benchmark1_raw_data}
    \centering
    \begin{tabular}{l | r r r r | r r r r }
    \toprule
        & \multicolumn{4}{c|}{2AFC score} & \multicolumn{4}{c}{JND score} \\
        Architecture & E & S & M & L & E & S & M & L \\\hline
        \multicolumn{1}{c|}{\textit{VGG Networks}} \\
        VGG-11 & 0.635 &	0.650 &	0.659 &	0.682 & 0.545 &	0.564 &	0.573 &	0.623 \\
        VGG-16 & 0.637 &	0.641 &	0.653 &	0.671 & 0.557 &	0.552 &	0.567 &	0.614 \\
        VGG-16\_bn & 0.629 &	0.636 &	0.642 &	0.669 & 0.545 &	0.545 &	0.556 &	0.621 \\
        VGG-19 & 0.635 &	0.639 &	0.653 &	0.666 & 0.557 &	0.547 &	0.568 &	0.608 \\
        \multicolumn{1}{c|}{\textit{Residual Networks}} \\
        ResNet-18 & 0.629 &	0.647 &	0.654 &	0.674 & 0.517 &	0.545 &	0.559 &	0.604 \\
        ResNet-50 & 0.629 &	0.661 &	0.658 &	0.659 & 0.525 &	0.561 &	0.573 &	0.593 \\
        ResNeXt-50 32x4d & 0.628 &	0.639 &	0.652 &	0.658 & 0.526 &	0.540 &	0.558 &	0.589 \\
        \multicolumn{1}{c|}{\textit{Inception Networks}} \\
        GoogLeNet & 0.631 &	0.677 &	0.682 &	0.667 & 0.512 &	0.542 &	0.563 &	0.567 \\
        InceptionNet v3 & 0.625 &	0.669 &	0.659 &	0.660 & 0.516 &	0.572 &	0.559 &	0.555 \\
        \multicolumn{1}{c|}{\textit{EfficientNet}} \\
        EfficientNet\_B0 & 0.626 &	0.621 &	0.665 &	0.651 & 0.530 &	0.540 &	0.598 &	0.587 \\
        EfficientNet\_B7 & 0.629 &	0.624 &	0.658 &	0.638 & 0.513 &	0.551 &	0.579 &	0.536 \\
        \multicolumn{1}{c|}{\textit{Uncategorized Networks}} \\ 
        AlexNet & 0.643 &	0.686 &	0.691 &	0.688 & 0.518 &	0.569 &	0.586 &	0.589 \\
        DenseNet-121 & 0.628 &	0.648 &	0.668 &	0.661 & 0.506 &	0.544 &	0.576 &	0.579 \\
        SqueezeNet & 0.635 &	0.679 &	0.685 &	0.678 & 0.527 &	0.573 &	0.602 &	0.606 \\
    \bottomrule
    \end{tabular}
\end{table*}

\begin{table*}[ht]
    \caption{PSNR of $\times4$ upscaling on Set~5, Set~14, and BSD100 of an Image Transformation Network trained with different loss networks}
    \label{tab:benchmark2_raw_data_4x_psnr}
    \centering
    \begin{tabular}{l | r r r r | r r r r | r r r r}
    \toprule
        & \multicolumn{4}{c|}{Set 5} & \multicolumn{4}{c}{Set 14} & \multicolumn{4}{c}{BSD100} \\
        Architecture & E & S & M & L & E & S & M & L & E & S & M & L \\\hline
        \multicolumn{1}{c|}{\textit{VGG Networks}} \\
        VGG-11 &  30.7 &	28.6 &	28.1 &	27.2 & 26.4 &	24.9 &	24.7 &	24.7 & 26.3 &	24.9 &	24.7 &	25.2 \\
        VGG-16 &  30.3 &	27.2 &	24.9 &	25.5 & 26.3 &	24.7 &	23.7 &	24.1 & 25.9 &	24.9 &	23.8 &	23.9 \\
        VGG-16\_bn &  28.6 &	28.5 &	23.2 &	13.2 & 24.7 &	24.5 &	21.9 &	14.3 & 25.2 &	24.4 &	22.7 &	15.4 \\
        VGG-19 & 30.2 &	26.1 &	26.2 &	26.8 & 26.3 &	24.1 &	24.0 &	24.0 & 26.1 &	24.3 &	24.2 &	23.5 \\
        \multicolumn{1}{c|}{\textit{Residual Networks}} \\
        ResNet-18 & 28.7 &	26.8 &	28.3 &	25.5 & 25.4 &	24.1 &	24.8 &	23.1 & 25.9 &	24.5 &	24.6 &	23.6 \\
        ResNet-50 & 29.1 &	31.0 &	28.4 &	24.1 & 25.5 &	25.7 &	25.0 &	22.5 & 25.4 &	25.2 &	24.8 &	24.5 \\
        ResNeXt-50 32x4d & 29.8 &	27.3 &	29.5 &	24.7 & 25.9 &	24.7 &	25.1 &	22.4 & 26.0 &	25.1 &	25.3 &	22.7 \\
        \multicolumn{1}{c|}{\textit{Inception Networks}} \\
        GoogLeNet & 26.5 &	29.5 &	28.4 &	26.0 & 24.2 &	25.3 &	24.7 &	23.5 & 24.7 &	25.0 &	24.9 &	24.6 \\
        InceptionNet v3 & 27.4 &	27.4 &	26.3 &	24.9 & 24.2 &	24.7 &	24.0 &	23.7 & 24.5 &	25.4 &	24.0 &	24.0 \\
        \multicolumn{1}{c|}{\textit{EfficientNet}} \\
        EfficientNet\_B0 & 21.1 &	26.8 &	27.3 &	25.8 & 20.7 &	24.5 &	24.3 &	24.1 & 21.1 &	25.0 &	24.8 &	24.5 \\
        EfficientNet\_B7 & 28.6 &	27.7 &	27.9 &	26.1 & 25.5 &	24.8 &	24.6 &	23.4 & 25.6 &	25.8 &	24.9 &	23.5 \\
        \multicolumn{1}{c|}{\textit{Uncategorized Networks}} \\ 
        AlexNet & 29.9 &	29.2 &	30.0 &	29.7 & 25.7 &	24.9 &	25.2 &	25.2 & 25.7 &	25.1 &	24.9 &	24.9 \\
        DenseNet-121 & 24.8 &	30.1 &	29.3 &	25.1 & 23.4 &	25.2 &	25.2 &	22.4 & 24.0 &	24.9 &	24.9 &	24.5 \\
        SqueezeNet & 31.1 &	29.1 &	28.4 &	26.9 & 26.5 &	25.0 &	24.8 &	23.9 & 26.2 &	25.3 &	24.6 &	24.1 \\
    \bottomrule
    \end{tabular}
\end{table*}

\begin{table*}[ht]
    \caption{MSSIM of $\times4$ upscaling on Set~5, Set~14, and BSD100 of an Image Transformation Network trained with different loss networks}
    \label{tab:benchmark2_raw_data_4x_mssim}
    \centering
    \begin{tabular}{l | r r r r | r r r r | r r r r}
    \toprule
        & \multicolumn{4}{c|}{Set 5} & \multicolumn{4}{c}{Set 14} & \multicolumn{4}{c}{BSD100} \\
        Architecture & E & S & M & L & E & S & M & L & E & S & M & L \\\hline
        \multicolumn{1}{c|}{\textit{VGG Networks}} \\
        VGG-11 & 0.855 &	0.772 &	0.812 &	0.793 & 0.690 &	0.603 &	0.630 &	0.643 & 0.646 &	0.546 &	0.579 &	0.602 \\
        VGG-16 & 0.859 &	0.824 &	0.806 &	0.804 & 0.689 &	0.649 &	0.640 &	0.643 & 0.639 &	0.603 &	0.589 &	0.583 \\
        VGG-16\_bn & 0.834 &	0.832 &	0.807 &	0.382 & 0.664 &	0.660 &	0.653 &	0.318 & 0.613 &	0.604 &	0.614 &	0.357 \\
        VGG-19 & 0.856 &	0.809 &	0.815 &	0.803 & 0.693 &	0.641 &	0.639 &	0.625 & 0.641 &	0.587 &	0.591 &	0.551 \\
        \multicolumn{1}{c|}{\textit{Residual Networks}} \\
        ResNet-18 & 0.826 &	0.838 &	0.834 &	0.805 & 0.672 &	0.665 &	0.669 &	0.629 & 0.638 &	0.628 &	0.619 &	0.574 \\
        ResNet-50 & 0.835 &	0.856 &	0.837 &	0.742 & 0.671 &	0.677 &	0.668 &	0.619 & 0.621 &	0.623 &	0.617 &	0.618 \\
        ResNeXt-50 32x4d & 0.840 &	0.836 &	0.846 &	0.824 & 0.678 &	0.671 &	0.673 &	0.647 & 0.636 &	0.631 &	0.628 &	0.594 \\
        \multicolumn{1}{c|}{\textit{Inception Networks}} \\
        GoogLeNet & 0.822 &	0.847 &	0.832 &	0.773 & 0.655 &	0.670 &	0.651 &	0.615 & 0.621 &	0.618 &	0.598 &	0.592 \\
        InceptionNet v3 & 0.786 &	0.825 &	0.813 &	0.779 & 0.615 &	0.667 &	0.653 &	0.627 & 0.568 &	0.626 &	0.609 &	0.590 \\
        \multicolumn{1}{c|}{\textit{EfficientNet}} \\
        EfficientNet\_B0 & 0.749 &	0.815 &	0.826 &	0.808 & 0.617 &	0.668 &	0.663 &	0.649 & 0.587 &	0.637 &	0.622 &	0.606 \\
        EfficientNet\_B7 & 0.826 &	0.803 &	0.812 &	0.797 & 0.663 &	0.648 &	0.648 &	0.629 & 0.615 &	0.616 &	0.607 &	0.582 \\
        \multicolumn{1}{c|}{\textit{Uncategorized Networks}} \\ 
        AlexNet & 0.828 &	0.803 &	0.815 &	0.814 & 0.665 &	0.627 &	0.624 &	0.632 & 0.618 &	0.581 &	0.562 &	0.573 \\
        DenseNet-121 & 0.784 &	0.851 &	0.845 &	0.777 & 0.628 &	0.678 &	0.674 &	0.624 & 0.592 &	0.625 &	0.626 &	0.604 \\
        SqueezeNet & 0.861 &	0.816 &	0.831 &	0.808 & 0.693 &	0.641 &	0.655 &	0.637 & 0.646 &	0.596 &	0.605 &	0.582 \\
    \bottomrule
    \end{tabular}
\end{table*}

\begin{table*}[ht]
    \caption{PSNR of $\times8$ upscaling on Set~5, Set~14, and BSD100 of an Image Transformation Network trained with different loss networks}
    \label{tab:benchmark2_raw_data_8x_psnr}
    \centering
    \begin{tabular}{l | r r r r | r r r r | r r r r}
    \toprule
        & \multicolumn{4}{c|}{Set 5} & \multicolumn{4}{c}{Set 14} & \multicolumn{4}{c}{BSD100} \\
        Architecture & E & S & M & L & E & S & M & L & E & S & M & L \\\hline
        \multicolumn{1}{c|}{\textit{VGG Networks}} \\
        VGG-11 & 26.2 &	26.0 &	22.8 &	19.6 & 22.8 &	22.5 &	20.9 &	18.4 & 22.6 &	22.1 &	21.3 &	19.0 \\
        VGG-16 & 25.2 &	23.4 &	19.9 &	18.6 & 22.5 &	21.4 &	19.5 &	19.0 & 22.3 &	21.5 &	20.1 &	19.7 \\
        VGG-16\_bn & 24.6 &	21.6 &	13.2 &	12.0 & 21.6 &	20.0 &	13.0 &	11.9 & 21.3 &	20.1 &	13.9 &	12.8 \\
        VGG-19 & 25.4 &	22.5 &	19.7 &	20.0 & 22.6 &	21.0 &	19.3 &	19.1 & 22.5 &	21.3 &	20.0 &	19.4 \\
        \multicolumn{1}{c|}{\textit{Residual Networks}} \\
        ResNet-18 & 25.6 &	24.9 &	23.2 &	22.1 & 22.4 &	21.8 &	20.5 &	20.3 & 22.3 &	21.5 &	20.7 &	20.9 \\
        ResNet-50 & 24.7 &	25.3 &	25.1 &	23.3 & 21.9 &	21.7 &	21.6 &	21.0 & 22.1 &	21.3 &	21.3 &	21.3 \\
        ResNeXt-50 32x4d & 25.5 &	25.3 &	25.0 &	14.7 & 22.4 &	21.9 &	21.5 &	15.7 & 22.3 &	21.7 &	21.2 &	16.2 \\
        \multicolumn{1}{c|}{\textit{Inception Networks}} \\
        GoogLeNet & 22.0 &	25.7 &	25.2 &	24.1 & 20.9 &	21.9 &	21.7 &	21.3 & 21.2 &	21.5 &	21.5 &	21.1 \\
        InceptionNet v3 & 22.6 &	24.2 &	23.8 &	24.5 & 20.7 &	21.7 &	21.7 &	21.9 & 21.3 &	21.7 &	21.7 &	21.8 \\
        \multicolumn{1}{c|}{\textit{EfficientNet}} \\
        EfficientNet\_B0 & 18.4 &	24.4 &	25.4 &	24.8 & 18.8 &	22.0 &	22.2 &	22.0 & 19.3 &	22.2 &	21.9 &	21.8 \\
        EfficientNet\_B7 & 25.0 &	23.5 &	25.8 &	24.6 & 22.2 &	21.3 &	22.4 &	21.8 & 22.4 &	21.9 &	22.2 &	22.0 \\
        \multicolumn{1}{c|}{\textit{Uncategorized Networks}} \\ 
        AlexNet & 26.3 &	25.8 &	25.5 &	24.2 & 22.6 &	21.9 &	21.9 &	21.4 & 22.1 &	21.5 &	21.6 &	21.4 \\
        DenseNet-121 & 21.4 &	24.1 &	24.9 &	23.9 & 20.8 &	21.6 &	21.9 &	21.3 & 21.4 &	21.8 &	21.5 &	21.4 \\
        SqueezeNet & 26.5 &	25.0 &	23.8 &	22.1 & 22.9 &	22.0 &	21.5 &	20.6 & 22.6 &	21.7 &	21.0 &	20.4 \\
    \bottomrule
    \end{tabular}
\end{table*}

\begin{table*}[ht]
    \caption{MSSIM of $\times8$ upscaling on Set~5, Set~14, and BSD100 of an Image Transformation Network trained with different loss networks}
    \label{tab:benchmark2_raw_data_8x_mssim}
    \centering
    \begin{tabular}{l | r r r r | r r r r | r r r r}
    \toprule
        & \multicolumn{4}{c|}{Set 5} & \multicolumn{4}{c}{Set 14} & \multicolumn{4}{c}{BSD100} \\
        Architecture & E & S & M & L & E & S & M & L & E & S & M & L \\\hline
        \multicolumn{1}{c|}{\textit{VGG Networks}} \\
        VGG-11 & 0.709 &	0.675 &	0.590 &	0.604 & 0.487 &	0.452 &	0.389 &	0.413 & 0.452 &	0.391 &	0.354 &	0.377 \\
        VGG-16 & 0.673 &	0.666 &	0.605 &	0.569 & 0.463 &	0.453 &	0.417 &	0.420 & 0.420 &	0.413 &	0.390 &	0.394 \\
        VGG-16\_bn & 0.685 &	0.649 &	0.518 &	0.380 & 0.478 &	0.458 &	0.385 &	0.282 & 0.448 &	0.443 &	0.401 &	0.321 \\
        VGG-19 & 0.696 &	0.661 &	0.591 &	0.574 & 0.481 &	0.450 &	0.402 &	0.399 & 0.441 &	0.419 &	0.373 &	0.369 \\
        \multicolumn{1}{c|}{\textit{Residual Networks}} \\
        ResNet-18 & 0.685 &	0.657 &	0.673 &	0.532 & 0.480 &	0.446 &	0.465 &	0.374 & 0.449 &	0.392 &	0.424 &	0.357 \\
        ResNet-50 & 0.678 &	0.688 &	0.685 &	0.588 & 0.479 &	0.464 &	0.456 &	0.402 & 0.443 &	0.416 &	0.398 &	0.375 \\
        ResNeXt-50 32x4d & 0.696 &	0.695 &	0.690 &	0.341 & 0.483 &	0.476 &	0.468 &	0.272 & 0.449 &	0.438 &	0.413 &	0.290 \\
        \multicolumn{1}{c|}{\textit{Inception Networks}} \\
        GoogLeNet & 0.617 &	0.695 &	0.673 &	0.620 & 0.454 &	0.459 &	0.460 &	0.427 & 0.438 &	0.414 &	0.401 &	0.369 \\
        InceptionNet v3 & 0.634 &	0.661 &	0.656 &	0.628 & 0.451 &	0.448 &	0.462 &	0.437 & 0.420 &	0.400 &	0.431 &	0.409 \\
        \multicolumn{1}{c|}{\textit{EfficientNet}} \\
        EfficientNet\_B0 & 0.592 &	0.669 &	0.688 &	0.667 & 0.436 &	0.469 &	0.466 &	0.460 & 0.413 &	0.443 &	0.413 &	0.428 \\
        EfficientNet\_B7 & 0.674 &	0.644 &	0.662 &	0.599 & 0.470 &	0.447 &	0.467 &	0.409 & 0.442 &	0.433 &	0.423 &	0.382 \\
        \multicolumn{1}{c|}{\textit{Uncategorized Networks}} \\ 
        AlexNet & 0.681 &	0.636 &	0.631 &	0.611 & 0.469 &	0.413 &	0.412 &	0.406 & 0.425 &	0.366 &	0.364 &	0.364 \\
        DenseNet-121 & 0.605 &	0.648 &	0.650 &	0.582 & 0.438 &	0.459 &	0.436 &	0.398 & 0.428 &	0.419 &	0.388 &	0.364 \\
        SqueezeNet & 0.710 &	0.606 &	0.627 &	0.611 & 0.491 &	0.406 &	0.417 &	0.417 & 0.455 &	0.362 &	0.354 &	0.369 \\
    \bottomrule
    \end{tabular}
\end{table*}

\begin{table*}[ht]
    \caption{Correctness, Completeness, and Quality on the MRD testset for U-net models trained with different loss networks}
    \label{tab:benchmark3_raw_data}
    \centering
    \begin{tabular}{l | r r r r | r r r r | r r r r}
    \toprule
        & \multicolumn{4}{c|}{Correctness} & \multicolumn{4}{c}{Completeness} & \multicolumn{4}{c}{Quality} \\
        Architecture & E & S & M & L & E & S & M & L & E & S & M & L \\\hline
        \multicolumn{1}{c|}{\textit{VGG Networks}} \\
        VGG-11 & 31.0 &	28.6 &	22.9 &	41.7 &	45.4 &	42.5 &	34.7 &	64.0 &	17.6 &	16.3 &	13.0 &	23.7 \\
        VGG-16 & 41.3 &	31.5 &	38.2 &	34.7 &	63.5 &	45.9 &	57.4 &	50.2 &	23.5 &	17.9 &	21.7 &	19.7 \\
        VGG-16\_bn &  32.3 &	36.1 &	34.9 &	29.8 &	47.1 &	54.6 &	53.2 &	43.9 &	18.4 &	20.5 &	19.9 &	17.0 \\
        VGG-19 & 39.2 &	37.8 &	42.1 &	29.0 &	60.9 &	56.9 &	64.4 &	43.0 &	22.4 &	21.5 &	23.9 &	16.5 \\
        \multicolumn{1}{c|}{\textit{Residual Networks}} \\
        ResNet-18 & 27.7 &	33.0 &	35.7 &	26.5 &	41.3 &	48.1 &	54.2 &	39.7 &	15.8 &	18.8 &	20.3 &	15.1 \\
        ResNet-50 & 40.1 &	22.5 &	26.9 &	30.6 &	62.0 &	34.2 &	40.2 &	44.8 &	22.8 &	12.8 &	15.3 &	17.4 \\
        ResNeXt-50 32x4d & 36.5 &	23.3 &	40.5 &	33.5 &	55.1 &	35.3 &	62.4 &	48.6 &	20.7 &	13.3 &	23.1 &	19.0 \\
        \multicolumn{1}{c|}{\textit{Inception Networks}} \\
        GoogLeNet & 25.3 &	24.1 &	26.1 &	36.9 &	38.2 &	36.6 &	39.1 &	55.7 &	14.4 &	13.8 &	14.9 &	21.0 \\
        InceptionNet v3 & 25.7 &	24.9 &	38.6 &	40.9 &	38.7 &	37.5 &	57.8 &	62.9 &	14.7 &	14.2 &	21.9 &	23.3 \\
        \multicolumn{1}{c|}{\textit{EfficientNet}} \\
        EfficientNet\_B0 & 33.9 &	31.9 &	34.8 &	35.9 &	49.2 &	46.5 &	48.9 &	49.8 &	19.3 &	18.1 &	19.5 &	21.3 \\
        EfficientNet\_B7 & 23.7 &	39.0 &	41.3 &	43.4 &	35.9 &	58.3 &	60.9 &	61.8 &	13.5 &	22.2 &	24.0 &	25.6 \\
        \multicolumn{1}{c|}{\textit{Uncategorized Networks}} \\ 
        AlexNet & 27.3 &	28.1 &	32.7 &	37.3 &	40.7 &	41.8 &	47.5 &	56.3 &	15.5 &	16.0 &	18.6 &	21.2 \\
        DenseNet-121 & 34.3 &	34.3 &	30.2 &	35.3 &	49.7 & 33.7 & 44.3 &	53.7 &	19.5 &	12.6 &	17.2 &	20.1 \\
        SqueezeNet & 21.6 &	39.6 &	29.4 &	24.5 &	33.2 &	61.4 &	43.4 &	36.9 &	12.3 &	22.6 &	16.8 &	13.9 \\
    \bottomrule
    \end{tabular}
\end{table*}

\begin{table*}[ht]
    \caption{Test accuracy on SVHN and STL-10 for downstream models using autoencoder encodings trained with different loss networks}
    \label{tab:benchmark4_raw_data}
    \centering
    \begin{tabular}{l | r r r r | r r r r}
    \toprule
        & \multicolumn{4}{c|}{SVHN} & \multicolumn{4}{c}{STL-10} \\
        Architecture & E & S & M & L & E & S & M & L \\\hline
        \multicolumn{1}{c|}{\textit{VGG Networks}} \\
        VGG-11 &        0.826 & 0.832 & 0.860 & 0.825 & 0.381 & 0.453 &	0.567 & 0.724 \\
        VGG-16 &        0.826 & 0.833 & 0.887 & 0.837 & 0.413 & 0.512 & 0.592 & 0.741 \\
        VGG-16\_bn &    0.816 & 0.820 & 0.863 & 0.492 & 0.434 & 0.318 & 0.247 & 0.465 \\
        VGG-19 &        0.826 & 0.833 & 0.896 & 0.821 & 0.397 & 0.499 & 0.607 & 0.736 \\
        \multicolumn{1}{c|}{\textit{Residual Networks}} \\
        ResNet-18 &         0.820 & 0.832 & 0.867 & 0.471 & 0.422 & 0.473 & 0.415 & 0.465\\
        ResNet-50 &         0.821 & 0.833 & 0.862 & 0.571 & 0.413 & 0.554 & 0.413 & 0.480\\
        ResNeXt-50 32x4d &  0.819 & 0.830 & 0.864 & 0.259 & 0.419 & 0.383 & 0.428 & 0.403\\
        \multicolumn{1}{c|}{\textit{Inception Networks}} \\
        GoogLeNet & 0.814 &	0.868 &	0.849 &	0.595 &	0.373 &	0.474 &	0.540 &	0.593\\
        InceptionNet v3 & 0.803 &	0.846 &	0.828 &	0.366 &	0.301 &	0.475 &	0.528 &	0.354 \\
        \multicolumn{1}{c|}{\textit{EfficientNet}} \\
        EfficientNet\_B0 & 0.818 &	0.828 &	0.838 &	0.714 &	0.393 &	0.384 &	0.434 &	0.695 \\
        EfficientNet\_B7 & 0.807 &	0.821 &	0.851 &	0.445 &	0.418 &	0.277 &	0.648 &	0.385 \\
        \multicolumn{1}{c|}{\textit{Uncategorized Networks}} \\ 
        AlexNet & 0.818 &	0.817 &	0.799 &	0.744 &	0.387 &	0.596 &	0.640 &	0.669 \\
        DenseNet-121 & 0.758 &	0.831 &	0.862 &	0.264 &	0.442 &	0.488 &	0.565 &	0.578 \\
        SqueezeNet & 0.825 &	0.845 &	0.843 &	0.791 &	0.405 &	0.513 &	0.639 &	0.681 \\
    \bottomrule
    \end{tabular}
\end{table*}


 




\vfill

\end{document}